\documentclass[runningheads]{llncs}
\usepackage{subcaption}
\usepackage{threeparttable}
\usepackage{caption}
\usepackage{graphicx}
\usepackage{amsmath,amssymb}
\usepackage{booktabs}
\usepackage{multirow}
\usepackage{enumitem}
\usepackage{hyperref}
\usepackage{xcolor}
\usepackage{siunitx}
\usepackage{array}
\usepackage{longtable}
\usepackage{tabularx}
\usepackage{ragged2e}
\usepackage{adjustbox}
\usepackage{makecell}
\usepackage{pifont}
\usepackage{float}
\usepackage{placeins}
\usepackage{afterpage}
\newcolumntype{Y}{>{\RaggedRight\arraybackslash}X}
\definecolor{wincol}{RGB}{220,255,220}   
\definecolor{cowincol}{RGB}{210,235,255} 
\newcommand{\hlwin}[1]{\textbf{#1}}

\newcolumntype{L}[1]{>{\raggedright\arraybackslash}p{#1}} 
\renewcommand{\arraystretch}{1.05} 

\setlist{nosep,leftmargin=*}
\hypersetup{colorlinks=true, linkcolor=blue, citecolor=blue, urlcolor=blue}

\newcommand{\rocket}{\textsc{Rocket}}
\newcommand{\sax}{\textsc{Sax}}
\newcommand{\sfa}{\textsc{Sfa}}

\newcommand{\fusionthree}{\textsc{Fusion3}}

\newcommand{\fusionsr}{\texttt{F2\_SR}}
\newcommand{\fusionsfr}{\texttt{F2\_SFR}}
\newcommand{\fusionss}{\texttt{F2\_SS}}

\title{A Regime-Aware Fusion Framework for Time Series Classification}
\titlerunning{Regime-Aware Fusion for TSC}
\author{Honey Singh Chauhan\inst{1} \and Zahraa S. Abdallah\inst{1}}
\authorrunning{H.S. Chauhan and Z.S. Abdallah}
\institute{School of Engineering Mathematics and Technology\\
University of Bristol, Bristol, UK\\
\email{honeyandbros@gmail.com, zahraa.abdallah@bristol.ac.uk}}

\begin{document}
\maketitle
\begin{abstract}
Kernel-based methods such as \rocket{} are among the most effective default approaches for univariate time series classification (TSC), yet they do not perform equally well across all datasets. We revisit the long-standing intuition that different representations capture complementary structure and show that selectively fusing them can yield consistent improvements over \rocket{} on specific, systematically identifiable kinds of datasets.
We introduce Fusion-3 (F3), a lightweight framework that adaptively fuses \rocket{}, \sax{}, and \sfa{} representations. To understand when fusion helps, we cluster UCR datasets into six groups using meta-features capturing series length, spectral structure, roughness, and class imbalance, and treat these clusters as interpretable data-structure regimes. Our analysis shows that fusion typically outperforms strong baselines in regimes with structured variability or rich frequency content, while offering diminishing returns in highly irregular or outlier-heavy settings.
To support these findings, we combine three complementary analyses: nonparametric paired statistics across datasets, ablation studies isolating the roles of individual representations, and attribution via SHAP to identify which dataset properties predict fusion gains. Sample-level case studies further reveal the underlying mechanism: fusion primarily improves performance by rescuing specific errors, with adaptive increases in frequency-domain weighting precisely where corrections occur.
Using 5-fold cross-validation on the 113 UCR datasets, F3 yields small but consistent average improvements over \rocket{}, supported by frequentist and Bayesian evidence and accompanied by clearly identifiable failure cases. Our results show that selectively applied fusion provides dependable and interpretable extension to strong kernel-based methods, correcting their weaknesses precisely where the data support it.

\end{abstract}

\section{Introduction}
\label{sec:intro}
Time series data, sequential records of observations over time, are a cornerstone of modern data analysis. They are generated in vast quantities across nearly every field of human endeavor, from the continuous monitoring of patient vital signs in healthcare and the high-frequency fluctuations of financial markets to the sensor readings from industrial machinery and the environmental data tracking climate change. The ability to automatically analyze and extract meaningful patterns from these sequences is a critical capability that drives decision-making, powers predictive systems, and unlocks scientific insight. At the heart of this analytical challenge lies Time Series Classification (TSC), the task of assigning a categorical label to a time series based on its underlying temporal patterns.

\noindent Despite its conceptual simplicity, TSC presents formidable challenges rooted in the sheer diversity of the data it encompasses. The term "time series" itself is broader than it suggests, applying to any form of sequential data, not just observations recorded over time. This means that alongside classic examples like financial tickers or ECG heartbeats, the field includes datasets that are counterintuitive yet powerful, such as the Yoga dataset from the UCR archive, which classifies poses from coordinate sequences, or GunPoint, which identifies hand movements from video. The patterns that define a class can manifest in vastly different forms—from subtle trends and periodic cycles to abrupt spikes and intricate symbolic motifs.

\noindent This inherent diversity creates a fundamental representational challenge. The most salient features for classification might be hidden within the raw data and can only be revealed by transforming the series into a different representation—a new format that highlights specific characteristics. For instance, a symbolic representation might expose recurring patterns, while a frequency representation could uncover underlying periodicities. Because no single representation is universally optimal, a method effective for one domain may be entirely unsuitable for another. Consequently, the development of robust, accurate, and general-purpose TSC algorithms that can navigate this representational landscape remains an active and vital area of research. Classical approaches range from distance-based methods (e.g., DTW nearest neighbour) and feature- or shapelet-based models to more recent deep architectures (CNNs, RNNs, Transformers) tailored to sequential data. Across this spectrum, no single approach consistently dominates, reflecting the diversity of TSC datasets in length, noise, and spectral structure. 

\noindent Within this landscape, kernel-based methods such as \rocket{} have emerged as lightweight, strong, and competitively accurate baselines for TSC, combining near state-of-the-art performance with very fast training and inference. However, TSC datasets differ dramatically in series length, spectral structure, roughness, and class imbalance, so no single representation---including \rocket{}---performs best everywhere. These differences induce distinct \emph{data-structure regimes}: for example, short spiky sequences, long smooth but frequency-structured signals, or highly imbalanced problems with weak class separation. This brings us back to a long-standing intuition: different representations specialise in different aspects of structure (e.g., convolutional kernels for local shapes, symbolic methods such as \sax{} for coarse shape and regime changes, and spectral methods such as \sfa{} for frequency content). The two questions we address in this paper are therefore: (i) how can we systematically discover the structure of a TSC dataset---that is, identify its regime---using meta-features; and (ii) how can we exploit this regime information by combining complementary representations through a simple, deployable fusion mechanism?

\noindent We answer these questions with a regime-aware framework that pairs meta-feature–based regime discovery with lightweight gated fusion of \rocket{}, \sax{}, and \sfa{} representations. This framework is supported by robust paired-comparison statistics, attribution analyses, and sample-level diagnostics, yielding a single, actionable recipe for practitioners: use fusion when meta-features indicate frequency complexity or long, structured series; otherwise, \rocket{} alone is sufficient.

\noindent This paper combines structure finding (regimes), fusion of representation and deep insights with case study and attribution methods into a single, actionable framework.

\noindent We summarise the contribution as follows: 
\begin{itemize}
    \item \textbf{Regime discovery.} We compute meta-features and cluster datasets into interpretable regimes (\emph{HighImb}, \emph{LongFSTCx}, \emph{SmoothSep}, \emph{HighFlCx}, \emph{HighCompOut}, \emph{ShortBase}), revealing actionable structure behind cross-dataset variability.
    \item \textbf{Lightweight fusion.} We introduce a gated neural architecture that combines \rocket{}, \sax{}, and \sfa{} embeddings (F3, a three-way fusion), plug-and-play on top of fast baselines.
    \item \textbf{Ablation studies.} Two-way fusions (F2: SAX+ROCKET, SFA+ROCKET) reveal which representation pairs matter in different regimes; SAX+SFA without ROCKET fails, confirming the convolutional backbone is essential.
    \item \textbf{Attribution \& case studies.} Global SHAP links gains to spectral complexity and series length; sample-level analyses expose \emph{rescued} vs.\ \emph{hurt} examples, confusion-matrix deltas, and regime-dependent gate weights (e.g., increased \sfa{} in frequency-structured settings).
    \item \textbf{Practical guidance.} Use F3 (three-way fusion: SAX+SFA+ROCKET) for regimes \emph{HighImb}, \emph{SmoothSep}, and \emph{HighFlCx}; use \fusionsr{} (two-way fusion: SAX+ROCKET) for regime \emph{ShortBase}; otherwise \rocket{} alone is a strong default.
\end{itemize}

\noindent\textbf{Paper organisation.}
The remainder of this paper is organised as follows. Section~\ref{sec:related} reviews related work on time series representations and their complementary strengths. Section~\ref{sec:methods} describes our method: the rationale for selecting SAX, SFA, and ROCKET, and the F3 gated fusion architecture. Section~\ref{sec:experiments} details the experimental setup: hyperparameter search strategy and meta-feature extraction for regime discovery. Section~\ref{sec:results} presents the main results in three stages: (i) regime discovery—six interpretable clusters capturing dataset structure; (ii) fusion performance—overall and per-regime comparisons via robust statistics, regime heatmap analysis; (iii) mechanistic insight—SHAP-based attribution linking meta-features to gains, case studies demonstrating how fusion helps, and ablation studies with two-way fusions. Section~\ref{sec:conclusion} concludes with practical recommendations, limitations, and future directions.

\section{Related Work}
\label{sec:related}
Using raw sequences with generic distances (e.g., Euclidean) is brittle under noise, scaling, and time warping \cite{Lin2007SAX}. A central design choice in TSC is therefore the \emph{representation}: a transformation that exposes structure useful for discrimination. Large empirical studies—most notably the ``Great Time Series Classification Bake Off'' \cite{Bagnall2017BakeOff} and its recent follow-up \cite{Middlehurst2021BakeoffRedux}—show that no single representation dominates across datasets, motivating families of complementary transformations.

\noindent The TSC literature offers a rich taxonomy of representations. \textbf{Shapelets} identify discriminative local subsequences via information gain or distance-based scoring \cite{Hills2014}. \textbf{Catch22} provides a compact set of 22 canonical time series features selected from over 7000 candidates for broad domain coverage \cite{Lubba2019Catch22}. \textbf{Continuous Wavelet Transform (CWT)} decomposes signals into time-frequency representations, useful for non-stationary patterns. \textbf{MultiROCKET} extends ROCKET with additional pooling statistics and multi-resolution kernels \cite{Tan2022MultiROCKET}. Dictionary-based methods (e.g., BOSS, cBOSS) combine symbolic discretisation with bag-of-patterns classifiers \cite{Schafer2015BOSS}. Deep learning approaches—CNNs, ResNets, InceptionTime, and Transformers—learn end-to-end hierarchical features but typically require more data and compute \cite{Fawaz2019Review,Ismail2020InceptionTime}.

\noindent We focus on three complementary representations: \textbf{SAX} (symbolic time-domain), \textbf{SFA} (symbolic frequency-domain), and \textbf{ROCKET} (random convolutional kernels). This choice is justified in Section~\ref{sec:methods} based on (i) domain diversity (time vs.\ frequency), (ii) empirical low correlation in large-scale benchmarks, and (iii) computational efficiency.

\noindent \textbf{SAX} maps $z$-normalised segment means (PAA) to symbols via Gaussian breakpoints, enabling lower bounds and motif discovery. Its behaviour depends on the windowing and alphabet parameters.
\textbf{SFA} instead truncates local DFTs and discretises each coefficient via Multiple Coefficient Binning, capturing global/spectral regularities and shift tolerance, often complementary to time-domain cues \cite{Schafer2012SFA}.
\textbf{ROCKET} replaces learned CNN filters with thousands of \emph{random} kernels of varied lengths/dilations and summarises each response by max and PPV; a linear/ridge classifier then operates on these features \cite{Dempster2020ROCKET}. The result is near-state-of-the-art accuracy with excellent speed and scalability. MiniROCKET \cite{Dempster2021MiniROCKET} and MultiROCKET \cite{Tan2022MultiROCKET} refine this recipe, but keep the same principle: diverse random convolutions + cheap pooling.

\noindent Because dataset characteristics vary, a long-standing question is how to identify the most discriminative representation \cite{Abanda2009}. Shapelet methods target local, discriminative motifs and perform well on some domains (e.g., ECG/outline datasets), but broad evaluations show no single approach dominates across datasets \cite{Middlehurst2021BakeoffRedux,Bagnall2017BakeOff}. Prior work has compared transformations via: (i) \emph{distance fidelity} (e.g., TLB for lower-bounded DTW surrogates) \cite{Wang2013TLB}; and (ii) \emph{global statistical criteria} (e.g., information gain, $F$-tests, Kruskal–Wallis) \cite{Hills2014}. Empirical studies also emphasise that performance hinges more on features than on the downstream classifier \cite{Nakano2014}.

\noindent Most comparisons operate at the \emph{dataset-level}. High TLB or superior average accuracy does not explain \emph{when} and \emph{why} a representation helps at the \emph{instance} level, nor how to adapt across heterogeneous data within a dataset \cite{Abanda2009,Serra2015Similarity}. Prior work has demonstrated the benefits of representation cooperation through ensemble methods that combine complementary symbolic representations, such as Co-eye \cite{Abdallah2020Coeye}, which uses multiple hyper-parameterised symbolic representations (PAA and Fourier approximations) with soft dynamic voting. However, non-adaptive ensembles aggregate evidence but rarely \emph{selectively} prioritise the most informative representation per sample in a way that is both effective and interpretable.

\noindent In this work, we study when symbolic (SAX/SFA) and convolutional (ROCKET) evidence is complementary and propose a simple, \emph{interpretable fusion} that adaptively re-weights representations per instance. Our analysis links \emph{meta-feature-derived clusters} (``regimes'') to systematic gains over ROCKET with robust statistics (Hodges–Lehmann medians, Wilcoxon, and Bayesian ROPE). At the sample level, case studies reveal \emph{which individual samples} benefit from fusion via confusion-matrix deltas, and learned fusion gate weights, connecting dataset regimes $\rightarrow$ representation utility $\rightarrow$ mechanistic understanding of when and why fusion helps.

\section{Method}
\label{sec:methods}

\subsection{Representations and Fusion}
We use \sax{} (symbolic time-domain), \sfa{} (symbolic frequency-domain), and \rocket{} (random convolutional kernels). This choice is justified by both theoretical and empirical evidence. 
Primarily, our choice spans complementary domains and resolutions. \textbf{SFA} operates in the frequency domain via truncated DFT and coefficient binning, exposing spectral periodicities and shift-invariant global structure. \textbf{SAX} and \textbf{ROCKET} both operate in the time domain, but capture \emph{fundamentally different aspects} of temporal structure: SAX provides coarse, noise-resistant symbolic summaries of segment-level trends (e.g., ``rising then flat"), whereas ROCKET extracts fine-grained, high-resolution local shapes and transients via thousands of random convolutional kernels (e.g., specific spike patterns, edge responses). This multi-resolution temporal coverage is a deliberate design choice validated by state-of-the-art ensemble methods—\textbf{HIVE-COTE} \cite{Lines2018HIVECOTE} and \textbf{Mr-Hydra} \cite{Dempster2023Hydra} explicitly combine multiple time-domain representations like shapelet/dictionary-based methods (akin to SAX's symbolic coarseness) with convolutional features (akin to ROCKET's fine-grained kernels) to achieve top performance. Empirically, the correlation matrix of accuracy ranks from benchmark studies \cite{Bagnall2017BakeOff,Middlehurst2021BakeoffRedux} confirms that SAX-like and ROCKET-like classifiers are often negatively correlated or weakly correlated, indicating they excel on different subsets of datasets.

\begin{figure}[htbp]
\centering
\includegraphics[width=1\linewidth]{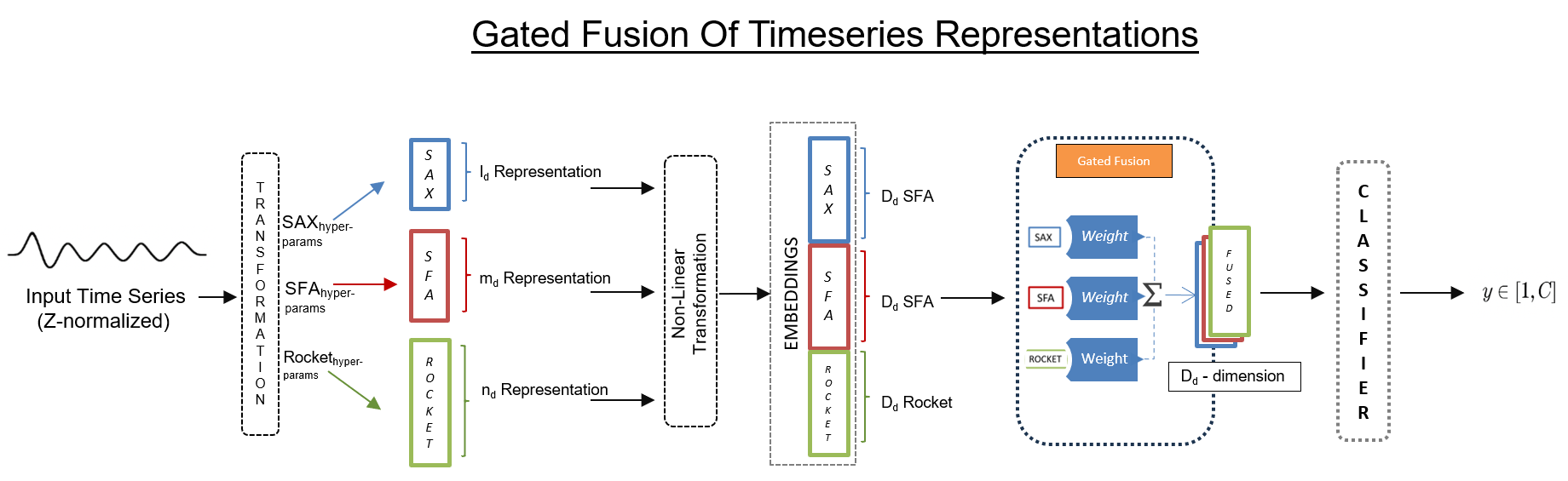}
\caption{Architecture of F3 (adaptive gated fusion). Parallel extraction of SAX, SFA, and ROCKET representations $\rightarrow$ dense embedding projection $\rightarrow$ instance-wise gating $\rightarrow$ weighted fusion $\rightarrow$ classification.}
\label{fig:model_architecture}
\end{figure}

\noindent \fusionthree{} (a three-way fusion of SAX, SFA, and ROCKET; hereafter ``F3") is a lightweight gated neural architecture that adaptively combines complementary time series representations. The complete workflow (Figure~\ref{fig:model_architecture}) proceeds in five stages: (1) Given an input time series, we extract three complementary representations in parallel—SAX produces symbolic time-domain features (dimensionality $\sim$~4000), SFA produces symbolic frequency-domain features (similar dimensionality), and ROCKET generates convolutional kernel features (again similar dimensionality from max/PPV pooling). (2) Each sparse representation is independently projected into a dense, lower-dimensional embedding space ($\mathbf{e}_{\text{SAX}}, \mathbf{e}_{\text{SFA}}, \mathbf{e}_{\text{ROCKET}} \in \mathbb{R}^d$, where $d\in\{64,128\}$) via fully connected layer + ReLU activation. (3) A gating network takes the concatenated embeddings and learns instance-specific importance weights $g=(g_{\sax},g_{\sfa},g_{\rocket})$, where $g_{\cdot}\!\in[0,1]$ and $\sum g_{\cdot}=1$, i.e. sigmoid activation followed by normalisation (ensuring $\sum g_{\cdot}=1$), enabling adaptive, interpretable, sample-level prioritisation. (4) The three embeddings are element-wise weighted by their gate values and summed: $\mathbf{e}_{\text{fused}} = g_{\sax} \cdot \mathbf{e}_{\text{SAX}} + g_{\sfa} \cdot \mathbf{e}_{\text{SFA}} + g_{\rocket} \cdot \mathbf{e}_{\text{ROCKET}}$. (5) The fused embedding is passed through a small MLP classifier (one hidden layer with dropout) to produce class logits, trained with cross-entropy loss via Adam. The learned gate weights provide post-hoc interpretability.

\noindent Furthermore, to understand which representation pairs contribute most to the observed gains in three-way fusion (F3), we also conduct ablation studies with two-way fusion variants (hereafter ``F2"): \fusionsfr{} (SFA+ROCKET), \fusionsr{} (SAX+ROCKET), and \fusionss{} (SAX+SFA), evaluated in Section~\ref{sec:ablation}.

\section{Experimental Setup}
\label{sec:experiments}

\subsection{Hyperparameter Search and Fairness (Full Grid)}
To ensure comparability, we used a \emph{full grid} over small, literature-backed \cite{Middlehurst2021BakeoffRedux} ranges and applied the \emph{same head capacity and training schedule} to all models. \textbf{SAX:} word $\{6,8\}$, frame $\{10,15,20\}$, alphabet $4$; \textbf{SFA:} word $\{6,8\}$, window $\{10,15,20\}$, alphabet $4$; \textbf{ROCKET:} $n_{\text{kernels}}\in\{1500,2000\}$, seed $=42$; \textbf{Head (all models):} embed/hidden $\{64,128\}$, dropout $0.2$; \textbf{Training:} LR $10^{-3}$, batch $32$, max $25$ epochs with patience $5$, $k{=}5$ folds, seed $42$. These ranges span coarse$\leftrightarrow$fine granularity in both time and frequency while matching prior studies \cite{Bagnall2017BakeOff,Middlehurst2021BakeoffRedux} and keeping cost tractable. Pilots showed broader ranges yielded $<1$\,pp median gain but $2$--$3\times$ runtime. Fixing the head and schedule prevents capacity/training time from confounding representation comparisons. Search depth: F3 ($6{\times}6{\times}2{\times}4=288$ configs/dataset), \fusionsr{}/\fusionsfr{} ($6{\times}2{\times}4=48$), SAX/SFA solo ($6{\times}4=24$), ROCKET solo ($2{\times}4=8$). Using a full grid ensures each model family receives the same search depth; training over HPC infrastructure made this feasible on all 113 datasets.

\subsection{Meta-Features and Regimes}
\label{sec:metafeat}
We compute a range of 13 low-correlated dataset meta-features categorised into two groups: \textbf{Global complexity features:} series length, turning points and variance, spectral entropy and its variance,
KL divergence of the power spectrum, permutation entropy, autocorrelation lag-1 and kurtosis. \textbf{Class separability features:} DTW separability time and frequency domain, Kruskal power spectral density of classes and imbalance index.
This grouping reflects two complementary logics. Global features describe the intrinsic properties of the time series themselves (e.g., entropy, length), which are useful for identifying broad structural similarities
across datasets. Class-based metrics, in contrast, explicitly exploit label information to measure separability (e.g., DTW class distances). Including both ensures that clustering is informed by how datasets
look in general, as well as how hard they are to separate in practice.
Feature choice was guided by prior work and practical utility. DTW-based separability follows Wang et al.~\cite{Wang2013TLB}, where lower-bound distances (LBKeogh) were shown as strong quality indicators. Statistical measures such as Kruskal PSD align with Hills et al.~\cite{Hills2014}, which emphasised distributional/statistical descriptors. Dataset-level factors like imbalance are well known to affect classification difficulty. Variance-based counterparts (e.g., turning\_points\_var, kurtosis\_var, spectral\_entropy\_var) were included to capture intra-dataset volatility. While averages capture central tendencies, variance reflects whether a property is consistent across all series or dominated by a few irregular ones. Capturing both aspects provides a richer description of the dataset structure.
Taken together, the final feature set spans global, class-based, and variance-sensitive perspectives, providing a balanced and logically grounded foundation for clustering. Detailed mathematical definitions and formulas for all meta-features are provided in Appendix C, Tables~\ref{tab:meta_features_compact} and~\ref{tab:meta_features_justification}.
Hierarchical clustering yields six regimes: \textbf{C1 HighImb}(high imbalance signals), \textbf{C2 LongFSTCx} (long, frequency-separable time-complex signals),
\textbf{C3 SmoothSep}(smooth and separable signals), \textbf{C4 HighFlCx} (highly fluctuating complex signals), \textbf{C5 HighCompOut}(high complexity outlier rich signals), \textbf{C6 ShortBase}(short baseline signals).

\noindent\textbf{Terminology.}
We use ``cluster'' to denote the unsupervised groups returned by Hierarchical Agglomerative Clustering on meta-features.
We use ``regime'' to denote the interpretable family of datasets characterised by those clusters (e.g., C2 LongFSTCx).

\section{Results}
\label{sec:results}

\noindent We evaluate on 113 UCR benchmark datasets (missing-value datasets excluded) spanning diverse domains, lengths, and class structures, using 5-fold cross-validation with hyperparameter grid search per model. All time series are $z$-normalised per instance (85 pre-normalised, 28 normalised before feature extraction). For each dataset, we form paired accuracy differences $\Delta_i = \text{acc}_i(\text{Fusion}) - \text{acc}_i(\text{ROCKET})$ and report four complementary statistical signals: (1) \textbf{HL-median $\Delta$pp + 95\% CI} (Hodges--Lehmann robust typical gain via Walsh averages, bootstrap CI); (2) \textbf{Wilcoxon signed-rank $p$} (two-sided, ties removed, Holm-adjusted across regimes); (3) \textbf{Bayesian $P(d{>}0)$} (posterior probability of improvement on a new dataset, Beta($\frac{1}{2},\frac{1}{2}$) prior on win/loss ratio); (4) \textbf{ROPE-$P_{\text{better}}$} (practical significance with data-dependent threshold $\delta_i=0.03(1-\text{acc}_{\text{ROCKET},i})$ clamped in $[0.10,2.0]$\,pp, measuring gains exceeding 3\% of baseline error).

\noindent Our findings address the two questions posed in the introduction. First, meta-feature clustering (Figure~\ref{fig:clustering}) reveals six interpretable regimes (Table~\ref{tab:regimes}), demonstrating systematic, discoverable structure in TSC data. Second (\S\ref{sec:overall}), fusion delivers statistically significant, regime-specific gains, with F3 (three-way fusion: SAX+SFA+ROCKET) winning overall and in three key regimes. We then examine \emph{why} fusion helps through SHAP attribution (\S\ref{sec:shap}) and \emph{how} it corrects errors through case studies (\S\ref{sec:case}), showing that frequency complexity predicts gains and gates adaptively upweight SFA where corrections occur.

\subsection{Clustering}
\label{sec:clustering}
\begin{figure}[htbp]
\centering
  \includegraphics[width=1\linewidth]{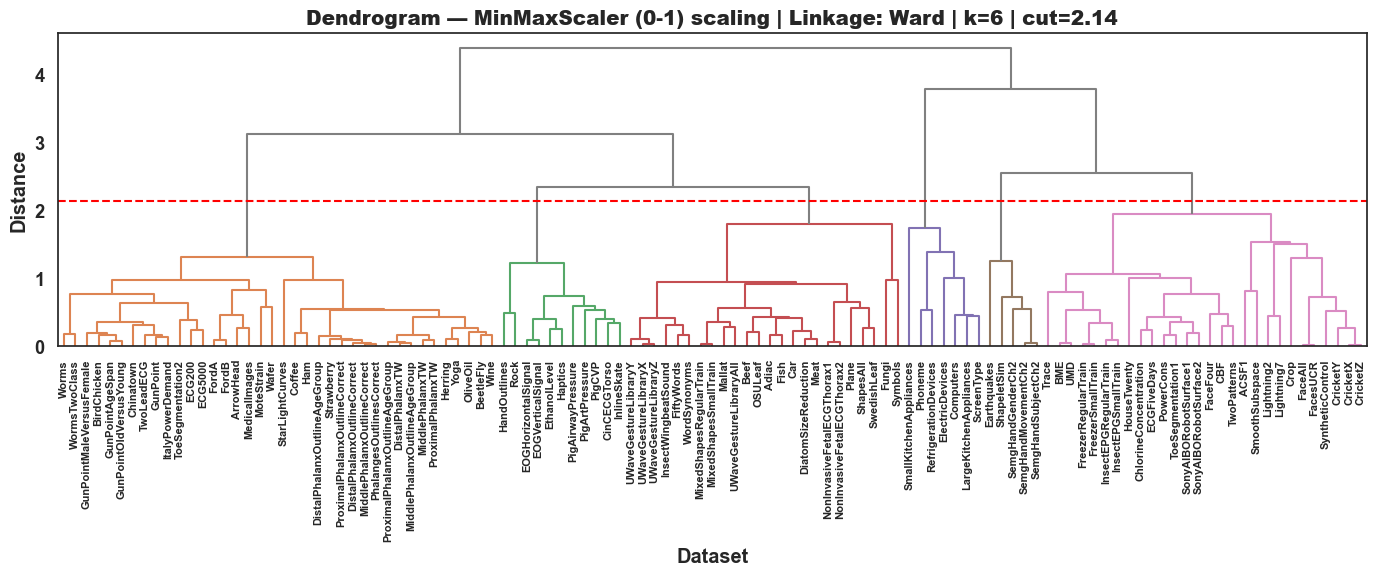}
  \caption{Dendrogram showing the hierarchical clustering of the 113 UCR/UEA datasets based on their 13 selected meta-features. The horizontal cut-off line indicates the division into six clusters, which are color-coded for clarity.}
  \label{fig:clustering}
\end{figure}

\noindent Hierarchical clustering on 13 meta-features (§\ref{sec:metafeat}) reveals six interpretable regimes that capture systematic variation in dataset structure (Table~\ref{tab:regimes}). The 113 datasets from the UCR/UEA archive were grouped using Hierarchical Agglomerative Clustering on the 13 handcrafted meta-features (each meta-feature is further summarised in Appendix C, Table~\ref{tab:meta_features_compact}). The resulting hierarchical structure is visualised in the dendrogram in Figure~\ref{fig:clustering}. By analyzing the dendrogram and cutting the tree at a Ward-linkage threshold (i.e., a threshold on the increase in within-cluster sum of squares) yielded six interpretable clusters, each representing a different archetype of a TSC problem. All meta-features were standardised prior to clustering, since Ward heights are scale-dependent. For the interested reader, we have also placed t-SNE \& UMAP 2-D projections of meta-features showing the clusters in Appendix D.2 (Figure~\ref{fig:umap_tsne}). The distribution of dataset types (device, image, motion, sensor) across clusters is visualised in Appendix D.1, Figure~\ref{fig:cdtype_dist}, revealing domain-specific clustering patterns.
Each regime represents a distinct family of TSC problems with shared characteristics in series length, spectral complexity, class separability, and imbalance. Quantitative meta-feature values per regime are visualised in Figure~\ref{fig:heatmap} (heatmap meta-feature row).

\begin{table}[h]
\centering
\caption{Six discovered regimes and their defining characteristics. $N$ = number of datasets per regime.}
\label{tab:regimes}
\scriptsize
\begin{tabular}{llp{8.5cm}}
\toprule
Regime & $N$ & Key Characteristics \\
\midrule
C1: HighImb & 38 & Datasets with high class imbalance (\texttt{imbalance\_index}=0.55, highest across regimes; Figure~\ref{fig:heatmap}), short/simple series, and moderate DTW separability. Many minority-class problems fall here, where standard classifiers struggle with skewed distributions. \\
C2: LongFSTCx & 11 & Long series (\texttt{ts\_length}=1856, highest; Figure~\ref{fig:heatmap}) with structured, frequency-separable signals (\texttt{dtw\_separability\_freq}=1.60). These datasets have rich temporal patterns that benefit from representations capturing both coarse trends and fine-grained shapes. \\
C3: SmoothSep & 24 & Smooth trajectories with high class separability (\texttt{dtw\_{\allowbreak}separability\_time}=2.37, \texttt{dtw\_{\allowbreak}separability\_freq}=2.10; Figure~\ref{fig:heatmap}). Classes are well-separated, and the signals have low roughness (\texttt{turning\_points}=0.11, lowest)—ideal conditions for fusion to add value. \\
C4: HighFlCx & 7 & Highly fluctuating signals with complex frequency patterns (\texttt{spectral\_entropy}=2.91, \texttt{global\_kl\_psd}=0.94) with high roughness (\texttt{turning\_points}=0.60; Figure~\ref{fig:heatmap}). These are often device/sensor datasets (e.g., \emph{RefrigerationDevices}, \emph{ElectricDevices}) where signals switch between different states and have rich frequency content. \\
C5: HighCompOut & 5 & High complexity, outlier-rich datasets (\texttt{kurtosis}=46.45, \texttt{spectral\_entropy}=4.68; Figure~\ref{fig:heatmap}). These contain irregular patterns and extreme values that make classification difficult for all methods. \\
C6: ShortBase & 28 & Short series (\texttt{ts\_length}=340) with jagged patterns (\texttt{turning\_points}=0.52) and modest spectral structure. The brevity limits what frequency-domain methods can capture. \\
\bottomrule
\end{tabular}
\end{table}

\noindent Regimes exhibit moderate separation in PCA meta-feature space (silhouette = 0.25, DBI = 1.19), consistent with partially overlapping dataset characteristics. Importantly, clusters are robust to scaling (ARI = 0.70 between MinMax and StandardScaler) and show moderate resampling stability (bootstrap ARI = 0.60 ± 0.16), indicating reproducible but soft regime boundaries.

\noindent These six regimes provide interpretable structure that exists in the UCR (113) datasets; we can now examine whether fusion performance varies systematically across regimes.

\subsection{Overall and Per-Regime Performance}
\label{sec:overall}

\noindent Across all 113 datasets, F3 (three-way fusion: SAX+SFA+ROCKET) improves over ROCKET with high statistical confidence and practical significance (Table~\ref{tab:overall_acc}). However, performance varies substantially by regime (Table~\ref{tab:overall-clusters}): fusion shows strong gains in three regimes, suggestive improvements in two more, and negative/negligible benefit in one. This regime-dependent pattern supports the hypothesis that complementary representations provide value in specific, identifiable data contexts.

\begin{table}[t]
\centering
\caption{Overall accuracy across 113 datasets (5-fold CV per dataset).
Acc is mean$\pm$SD (\%). $\Delta$pp and $\Delta$SD are differences vs.\ ROCKET (R).
Win-rate = Wins/(Wins+Losses). Ablation studies with two-way fusions are reported in Section~\ref{sec:ablation}. Distribution of accuracy gain is presented in Appendix D.3, Figure~\ref{fig:accuracy_gains}.}
\label{tab:overall_acc}
\begin{tabular}{lcccccc}
\toprule
Model & Acc $\pm$ SD (\%) & $\Delta$pp & $\Delta$SD & Wins/Losses/Ties & Win-rate \\
\midrule
\textit{R (Baseline)}     & 91.47 $\pm$ 9.82          & --             & --             & --               & -- \\
\textbf{F3 (SAX+SFA+R)} & \textbf{91.98 $\pm$ 9.59} & \textbf{+0.51} & \textbf{-0.23} & \textbf{80/12/21} & \textbf{87.0\%} \\
\bottomrule
\end{tabular}
\end{table}

\noindent In Table~\ref{tab:overall_acc}, each entry aggregates over 113 datasets (5-fold CV per dataset) and reports \emph{mean}$\pm$\emph{SD} accuracy; wins/losses/ties are computed per dataset, comparing mean accuracies to \rocket{}. F3 achieves 80 wins, 12 losses, and 21 ties (87.0\% win-rate) with a mean accuracy of 91.98$\pm$9.59\% versus 91.47$\pm$9.82\% for \rocket{}. F3 also reduces fold variance ($\Delta$SD $= -0.23$), indicating improved cross-validation stability. Detailed per-regime accuracy statistics for all models (including solo SAX, SFA, and two-way fusions) are provided in Appendix A, Table~\ref{tab:appendix_all_models_by_cluster}, while per-dataset accuracies are available in Appendix B, Table~\ref{tab:dataset-accuracies-by-cluster}.

\noindent Table~\ref{tab:overall-clusters} breaks down performance by regime. \textbf{Metric definitions:} \textbf{(1) HL-median $\Delta$pp + 95\% CI:} Hodges--Lehmann estimator (Walsh averages) as a robust typical gain; bootstrap CI. \textbf{(2) Wilcoxon signed-rank $p$:} Two-sided; ties removed; Holm-adjusted across clusters. \textbf{(3) Bayesian $P(d{>}0)$:} Beta($\frac12,\frac12$) prior on wins vs.\ losses gives a posterior mean and 95\% credible interval—``probability of improvement on a new dataset.'' \textbf{(4) ROPE-$P_{\text{better}}$:} Practical significance via a per-dataset threshold $\delta_i=\rho(1-\text{acc}_{\text{ROCKET},i})$ in pp (clamped to $[0.10,2.0]$\,pp); we use $\rho=0.03$ (3\% of baseline error). \emph{Win-rate} is (wins + ties) / total vs.\ ROCKET. 

\noindent We label \textbf{F3} as a clear \emph{winner} in a regime only when all of the following hold:
\begin{itemize}\itemsep0.2em
  \item HL--median $\Delta$pp $>0$ \textbf{and} Holm--adjusted Wilcoxon $p<0.05$;
  \item ROPE--$P_{\text{better}}\ge 0.5$.
\end{itemize}
For small regimes (\(N\le 12\)) where intervals are wide, we report results as \emph{suggestive but underpowered} when HL--median $>0$ but Holm--adjusted $p>0.05$.
Otherwise, we report \emph{no clear winner}.

\begin{table}[t]
\centering
\caption{Overall and per-regime summary: \textbf{F3} vs.\ \textbf{R} (ROCKET).
HL-median differences in percentage points (pp).
\textbf{Bold} = Holm-adjusted $p<.05$ (per-regime).
$\blacktriangle$ marks ROPE $P_{\text{better}}\!\ge\!0.50$ (practical uplift), considering per-dataset threshold $\delta_i=\rho(1-\text{acc}_{\rocket,i})$ in pp (clamped in $[0.10,2.0]$\,pp) and $\rho=0.03$ (3\% of baseline error).}
\label{tab:overall-clusters}
\scriptsize
\begin{tabular}{l r c c c c}
\toprule
Regime & $N$ &
HL $\Delta$pp [95\% CI] & Wilcoxon $p_{\text{Holm}}$ &
$P(d{>}0)$ & ROPE-$P_{\text{better}}$ \\
\midrule
Overall & 113 & \hlwin{\textbf{0.43} [0.31, 0.57]} $\blacktriangle$ & $<\!10^{-4}$ & 0.87 & 0.55 \\
\midrule
C1 HighImb & 38 & \hlwin{\textbf{0.51} [0.36, 0.78]} $\blacktriangle$ & \textbf{$<\!10^{-4}$} & 0.92 & 0.62 \\
C2 LongFSTCx & 11 & 0.42 [0.07, 1.53] & 0.1934 & 0.77 & 0.44 \\
C3 SmoothSep & 24 & \hlwin{\textbf{0.58} [0.36, 0.86]} $\blacktriangle$ & \textbf{$<\!10^{-4}$} & 0.98 & 0.77 \\
C4 HighFlCx & 7 & 1.09 [-0.63, 2.87] & 0.4375 & 0.81 & 0.41 \\
C5 HighCompOut & 5 & -0.72 [-4.28, 0.05] & 0.4375 & 0.30 & 0.08 \\
C6 ShortBase & 28 & 0.16 [0.01, 0.43] & 0.0639 & 0.78 & 0.39 \\
\bottomrule
\end{tabular}
\end{table}

\begin{itemize}\itemsep0.3em
    \item \textbf{Overall improvement.} F3 achieves HL–median gain of 0.43\,pp [0.31, 0.57] with Wilcoxon $p<10^{-4}$, Bayesian $P(d{>}0)=0.87$, and ROPE–$P_{\text{better}}=0.55$, indicating consistent winning improvements across the benchmark.
    
    \item \textbf{Regime-level variation.} C1 (HighImb) and C3 (SmoothSep) show the strongest winning evidence: HL-median gains of 0.51 and 0.58\, pp respectively, both Holm $p<10^{-4}$, ROPE $P_{\text{better}}\ge 0.62$. C4 (HighFlCx) indicates the highest gain and shows the largest point estimate (1.09\, pp) with posterior probability $P(d{>}0)=0.81$, though intervals are wide due to small sample size ($n=7$; we discuss C4 further in SHAP section~\ref{sec:shap}).
    
    \item \textbf{Weak or negative effects in some regimes.} C5 (HighCompOut) shows negative point estimates, though results are indicative of lower performance, but we delay making a strong conclusion until further analysis in SHAP section~\ref{sec:shap} due to the small $n=5$. C6 (ShortBase) shows marginal gains (Holm $p=0.064$). C2 (LongFSTCx) shows positive point estimates, but Holm-adjusted tests are non-significant ($n=11$).
    
    \item \textbf{Baseline strength.} ROCKET remains a strong baseline across most regimes. The sub-pp average gap and regime-dependent variation suggest fusion provides value in specific contexts rather than uniformly.

     \item \textbf{Practical significance.} Gains are modest in pp but \emph{reliable}. Where ROPE–$P_{\text{better}} \allowbreak \ge 0.5$ (e.g., C1, C3), improvements are not only consistent but practically meaningful under data-dependent margins; elsewhere, high $P(d{>}0)$ with sub-ROPE probabilities indicates many small wins rather than large shifts.
\end{itemize}

\noindent The regime heatmap (§\ref{sec:heatmap}) situates these results in meta-feature space. We then examine which meta-features predict gains through SHAP attribution (§\ref{sec:shap}) and investigate the correction mechanism through sample-level case studies (§\ref{sec:case}).

\subsection{Regime Heatmap: Meta-Features, Solo Strength, and Fusion Behaviour}

\begin{figure}[t]
\centering
\includegraphics[width=\textwidth]{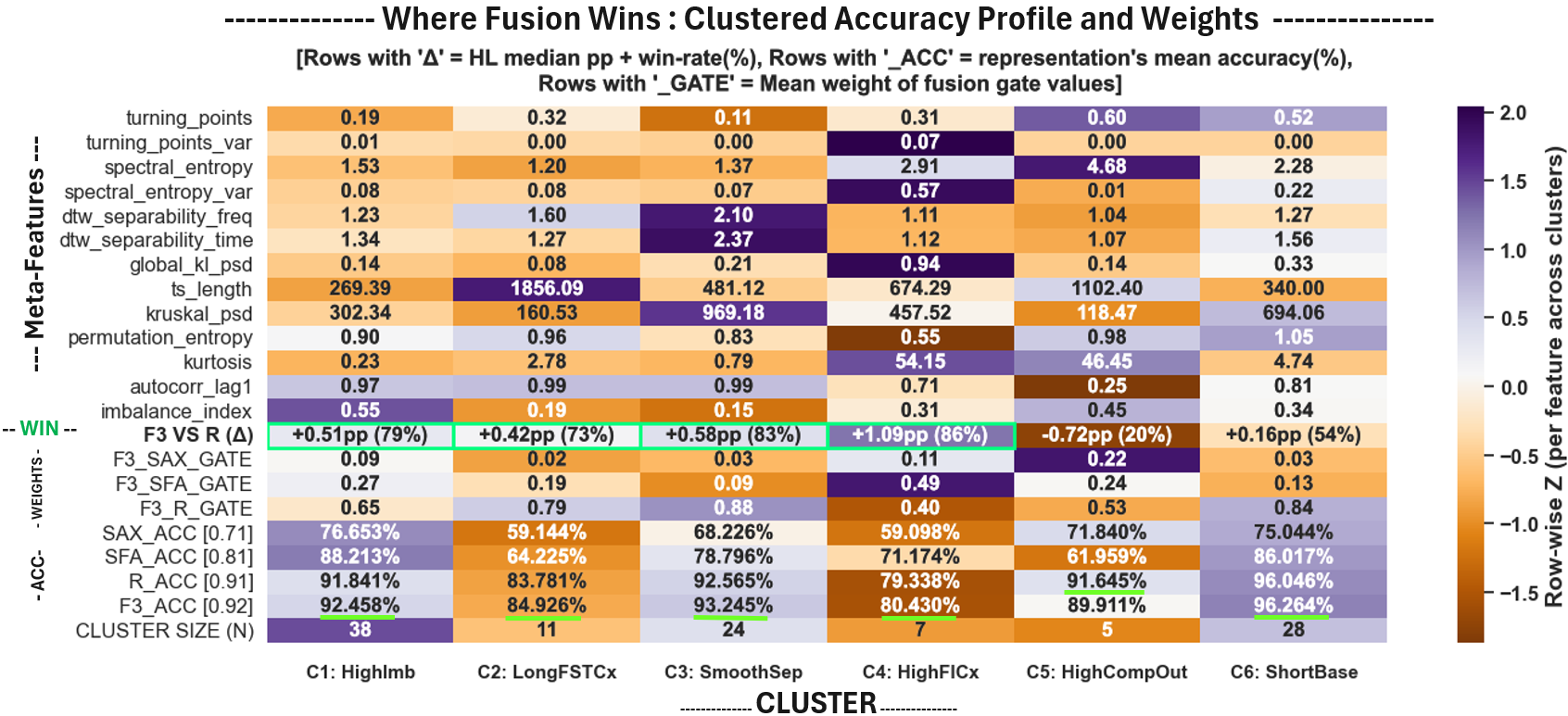}
\caption{
Regime heatmap summarising meta-features, solo accuracies, fusion gains, gate weights and cluster sizes.
Each row is standardised across the six regimes (row-wise Z-score), so colours indicate where a given quantity is relatively high or low \emph{within that row}; magnitudes should be read from the numeric annotations.
\textbf{Meta-features (top block):} entries are raw cluster means on the original scale.
\textbf{Fusion vs.\ ROCKET (middle block):} the ``F3 vs R (\(\Delta\))'' row reports the Hodges--Lehmann median accuracy difference in percentage points, annotated as ``\(\Delta\)pp (win-rate\%)''; green rectangles mark regime winners (and co-winners if they are within \(0.10\)pp in HL-median and the regime is small).
\textbf{Gate weights (lower-middle):} mean fusion gate weights for \fusionthree{} (SAX/SFA/ROCKET) within each regime.
\textbf{Solo accuracies (bottom block):} mean accuracies for SAX, SFA, ROCKET and \fusionthree{}.
The last row gives the number of datasets per regime.
Per-regime standard deviations for each model are reported in Appendix A, Table~\ref{tab:appendix_all_models_by_cluster}. A complementary visualisation showing normalized gate weight dominance across regimes is provided in Appendix D.3, Figure~\ref{fig:weight_dominance}.}
\label{fig:heatmap}
\end{figure}

\paragraph{How to read Fig.~\ref{fig:heatmap}.}
The figure is intended as the ``one-panel overview'' connecting the earlier paired tests (Table~\ref{tab:overall-clusters}) to the meta-feature space.
Each regime can be read along five aligned layers:
\textbf{(i)} what the data characteristics look like (meta-features),
\textbf{(ii)} how strong each solo representation is (SAX/SFA/ROCKET accuracies),
\textbf{(iii)} whether F3 improves over ROCKET on average and how often (HL $\Delta$pp + win-rate),
and \textbf{(iv)} where the fusion gate allocates weight across SAX, SFA and ROCKET, and \textbf{(v)} how many datasets support that conclusion.
This section therefore emphasises \emph{coherence across these layers}, while formal significance is handled by the paired tests (Table~\ref{tab:overall-clusters}).
Colours highlight relative patterns across regimes; the numbers carry the quantitative story.

\label{sec:heatmap}

\vspace{0.2em}

\paragraph{Regime summaries (what changes, what stays constant).}
Across regimes, ROCKET is generally the strongest solo representation, but the \emph{gap} to SFA and SAX varies markedly.
Where ROCKET and SFA are closer, F3 has more opportunity to improve by mixing frequency cues; where ROCKET dominates decisively, the gate concentrates on ROCKET, and F3 tends to yield smaller gains.
We now summarise each regime in this joint view, and explicitly point forward to the sections where we verify the mechanisms.

\vspace{0.4em}
\noindent\textbf{C1: HighImb (n=38)}
\emph{Meta-feature profile:} This is the largest regime (\(\approx 34\%\) of datasets).
It is characterised by strong class imbalance (imbalance index \(\approx 0.55\), high Z), short series on average (\(\texttt{ts\_length}\approx 270\)) and weak DTW separability (both time and frequency rows sit near the middle of the colour scale). The difficulty here is primarily dominated by label skew.
\emph{Solo strength:} ROCKET is best, but importantly, the ROCKET--SFA gap is \emph{smaller than the global average across all datasets}. This means SFA remains competitive on a non-trivial subset of datasets/samples even though ROCKET wins on average.
\emph{Fusion behaviour:} F3 shows a positive HL $\Delta$pp and a majority win-rate over ROCKET, consistent with ``many small corrections'' rather than a dramatic regime-level overhaul.
\emph{Gate behaviour:} The gate is ROCKET-heavy (reflecting ROCKET's solo advantage) but allocates a meaningful share to SFA (reflecting the reduced ROCKET--SFA separation), while SAX remains minor.
\emph{Interpretation:} In an imbalance-dominated regime, F3 behaves as a \emph{selective add-on}: it keeps ROCKET as the backbone and uses frequency cues to resolve borderline cases.
We revisit this mechanism at the sample level via rescued/hurt analyses in \S\ref{sec:case}.

\vspace{0.4em}
\noindent\textbf{C2: LongFSTCx (n=11)}
\emph{Meta-feature profile:} This regime contains longer time series with structured dynamics and non-trivial spectral texture (elevated frequency-domain separability).
\emph{Solo strength:} ROCKET exceedingly outperforms the other solo representations here, indicating that its random convolutional features already capture much of the discriminative structure.
\emph{Fusion behaviour:} F3 shows a positive point estimate in HL $\Delta$pp, but the regime is small, and the uncertainty is correspondingly large; this is the archetypal ``directionally consistent but underpowered'' setting.
\emph{Gate behaviour:} Consistent with solo performance, the gate remains concentrated on ROCKET, with SFA contributing intermittently rather than dominating.
\emph{Interpretation:} C2 is best treated as evidence about \emph{when fusion does not need to be aggressive}: when one representation is clearly strongest, fusion mostly preserves it.
We connect C2 to global attribution (length-related effects) in \S\ref{sec:shap} and to ablations comparing reduced fusion variants in \S\ref{sec:ablation}.

\vspace{0.4em}
\noindent\textbf{C3: SmoothSep (n=24)}
\emph{Meta-feature profile:} This regime exhibits comparatively clean separability signals (dark cells in the DTW-separability rows, moderate entropy), indicating that discriminative structure exists and is stable.
\emph{Solo strength:} Unsurprisingly, ROCKET is strong and remains the leading solo model with a significant margin over SFA and SAX.
\emph{Fusion behaviour:} F3 shows a clear positive HL $\Delta$pp with a strong win-rate, indicating that even when ROCKET is already strong, there is systematic room for improvement.
\emph{Gate behaviour:} The mean gate in this regime is strongly ROCKET-heavy (often the highest ROCKET weight among regimes), which is consistent with ROCKET being best solo. The key point is not that the gate shifts away from ROCKET, but that \emph{when the model does allocate weight to SFA/SAX, those allocations coincide with correctness more often than not} (validated at the sample level in \S\ref{sec:case}).
\emph{Interpretation:} C3 illustrates a common fusion pattern in this benchmark: the best behaviour is not ``replace ROCKET'', but ``keep ROCKET and fix what it misses''.

\vspace{0.4em}
\noindent\textbf{C4: HighFlCx (n=7)}
\emph{Meta-feature profile:} This small regime is characterised by high fluctuation/complexity in the frequency domain (e.g., high PSD-divergence and spectral variability), together with low \texttt{permutation\_entropy} and high \texttt{kurtosis}. Often associated with sensor/device datasets.
\emph{Solo strength:} ROCKET is weaker here than in most other regimes, and SFA tends to be relatively more competitive, shrinking the ROCKET--SFA gap compared to the global average.
\emph{Fusion behaviour:} F3 shows its largest regime-level point estimate (HL $\Delta$pp), but uncertainty is large because $n=7$.
\emph{Gate behaviour:} The gate shows a striking reallocation of mass towards SFA, making this the most SFA-dominated regime in the heatmap. This qualitatively matches the frequency-driven meta-feature signature.
\emph{Interpretation and flow control:} We deliberately avoid ``closing the loop'' here: C4 is where the heatmap provides a \emph{candidate mechanism} (frequency diversity $\rightarrow$ higher SFA weight $\rightarrow$ larger gains), but the correct place to validate this mechanism is global attribution (SHAP) and targeted case studies.
Accordingly, we return to C4 in \S\ref{sec:shap} (feature importance alignment) and \S\ref{sec:case} (dataset- and sample-level confusions and rescues).

\vspace{0.4em}
\noindent\textbf{C5: HighCompOut (n=5)}
\emph{Meta-feature profile:} This is the smallest regime and is dominated by complex/outlier-heavy structure, where variance and spikiness can distort both time- and frequency-domain summaries.
\emph{Solo strength and fusion behaviour:} Performance is variable, and uncertainty is large; F3 does not show a reliable advantage here and even exhibits negative point estimates.
\emph{Gate behaviour:} The gate remains ROCKET-dominant but, compared to other regimes, assigns relatively more weight to SAX, hinting at a regime where the fusion is less decisive than in ``easy'' regimes, reflecting instability rather than healthy adaptivity.
\emph{Interpretation:} C5 is best framed as a \emph{candidate failure regime}: small-$N$ prevents definitive conclusions, but it motivates why ablations and diagnostics matter.
We explicitly revisit this regime when discussing failure modes and simplified variants in \S\ref{sec:ablation} and \S\ref{sec:case}.

\vspace{0.4em}
\noindent\textbf{C6: ShortBase (n=28)}
\emph{Meta-feature profile:} This regime contains shorter series with jagged local structure, where time-domain roughness dominates, and frequency summaries are less stable.
\emph{Solo strength:} ROCKET remains strong and shows typical leads; SFA and SAX are weaker.
\emph{Fusion behaviour:} F3 improves only modestly (positive HL $\Delta$pp, but smaller than regimes where SFA is closer to ROCKET).
\emph{Gate behaviour:} The gate concentrates on ROCKET, consistent with solo dominance; contributions from other representations are comparatively small.
\emph{Interpretation:} C6 motivates why two-way fusions can be competitive when frequency structure is weak; we treat this explicitly in the ablation section (\S\ref{sec:ablation}), rather than overcrowding the heatmap.

\paragraph{Summary and handoff.}
The heatmap provides the \emph{context} that tables alone cannot: it shows that F3 gains occur where (a) ROCKET is not overwhelmingly superior to SFA/SAX, and/or (b) meta-features indicate frequency diversity or stable separability.
The next section (\S\ref{sec:shap}) tests this claim globally using attribution, and \S\ref{sec:case} then validates it at the dataset and sample level (confusions, rescued/hurt fractions, and gate shifts).


\section{Explaining When Fusion Corrects ROCKET via Meta-Feature Attribution}
\label{sec:shap}

The regime-level analysis in Section~\ref{sec:heatmap} indicates that the effectiveness of fusion varies substantially across datasets and appears closely tied to differences in meta-feature characteristics. To identify which dataset-level meta-features are most strongly associated with accuracy improvements of F3 over ROCKET, we use SHAP (SHapley Additive exPlanations) analysis.

For each dataset $d$, we consider the response
\begin{equation}
\Delta\mathrm{Acc}(d) = \mathrm{Acc}_{\text{F3}}(d) - \mathrm{Acc}_{\text{ROCKET}}(d),
\end{equation}
computed under the same 5-fold cross-validation protocol used throughout the paper. A regression model is trained to predict $\Delta\mathrm{Acc}$ from the dataset meta-features, and SHAP values are used to attribute the model’s predictions to individual features. Positive SHAP values indicate dataset properties associated with larger fusion gains, while negative values indicate conditions under which fusion is less effective.

\begin{figure}[t]
\centering
\includegraphics[width=\textwidth]{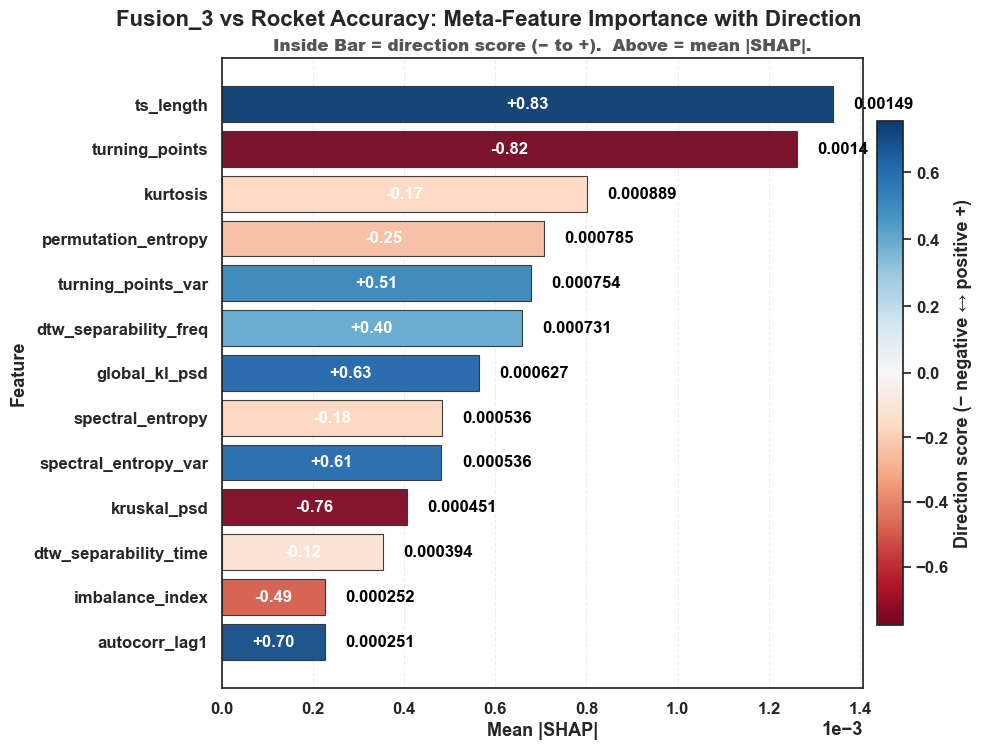}
\caption{SHAP summary for predicting $\Delta\!\operatorname{acc}=\operatorname{acc}(\fusionthree)-\operatorname{acc}(\rocket)$ from meta-features.
Higher mean absolute SHAP indicates stronger global influence; signs are taken from the SHAP expectation.}
\label{fig:shap}
\end{figure}

Figure~\ref{fig:shap} summarises the resulting attributions. Across datasets, F3 gains are most strongly associated with longer time series (\texttt{ts\_length}) and measures of spectral diversity such as \texttt{global\_kl\_psd} and \texttt{spectral\_entropy\_var}. These features characterize signals with rich and heterogeneous frequency structure, where a single representation may fail to capture all discriminative information.

In contrast, meta-features reflecting pervasive local irregularity—most notably \texttt{turning\_points} and \texttt{permutation\_entropy}—are negatively associated with fusion gains. A notable asymmetry emerges between these features and \texttt{turning\_points\_var}: while high overall irregularity suppresses gains, variability in local structure is positively associated with improved fusion performance. This suggests that fusion benefits from structured diversity in signal behaviour rather than uniformly high randomness.

Viewed through the lens of the previously identified regimes, datasets in the HighFlCx cluster tend to align very closely with the positively associated SHAP drivers, whereas datasets in the outlier-heavy HighCompOut cluster concentrate on negatively associated features. Therefore, these global attributions anticipate both the strongest practical gains observed in frequency-complex regimes (e.g., C4) and the consistent failure of fusion under extreme irregularity (C5).

Taken together, the SHAP analysis reinforces the view that F3 is most effective when frequency structure is diverse and informative, but degrades in settings dominated by local irregularity. The following section examines this behaviour more closely through two-way fusion ablations, clarifying how the individual components of F3 contribute to both its gains and its failure modes.


\section{Ablation Studies}
\label{sec:ablation}

The SHAP analysis in Section~\ref{sec:shap} identifies dataset-level properties associated with F3 gains but does not, by itself, explain how the individual representations contribute to these improvements or why fusion fails in certain settings. To clarify these mechanisms, understand which components are essential, and how different representations contribute under varying data conditions, we examine two-way fusion ablations: \fusionsfr{} (SFA+ROCKET), \fusionsr{} (SAX+ROCKET), and \fusionss{} (SAX+SFA without ROCKET). These variants remove one component at a time while retaining the same training and evaluation protocol, allowing us to attribute performance changes to the excluded representation. 

Table~\ref{tab:ablation_overall} summarises overall performance across all 113 datasets. Removing ROCKET (\fusionss{}) leads to a substantial performance collapse (-8.01 pp), confirming that convolutional features form the indispensable backbone of competitive performance. In contrast, both \fusionsfr{} and \fusionsr{} consistently outperform ROCKET, capturing a meaningful fraction of F3's overall gain (+0.16 pp and +0.13 pp, respectively), while remaining substantially simpler models.

\begin{table}[h]
\centering
\caption{Ablation: two-way fusions vs.\ ROCKET (R). Overall accuracy across 113 datasets (5-fold CV).
Acc is mean$\pm$SD (\%). $\Delta$pp and $\Delta$SD are differences vs.\ ROCKET.}
\label{tab:ablation_overall}
\begin{tabular}{lcccccc}
\toprule
Model & Acc $\pm$ SD (\%) & $\Delta$pp & $\Delta$SD & Wins/Losses/Ties & Win-rate \\
\midrule
\textit{R (Baseline)} & 91.47 $\pm$ 9.82 & -- & -- & -- & -- \\
F3 (SAX+SFA+R) & 91.98 $\pm$ 9.59 & +0.51 & -0.23 & 80/12/21 & 87.0\% \\
\midrule
\fusionsfr{} (SFA+R) & 91.63 $\pm$ 9.89 & +0.16 & +0.07 & 67/20/26 & 77.0\% \\
\fusionsr{} (SAX+R) & 91.60 $\pm$ 9.78 & +0.13 & -0.04 & 63/23/27 & 73.3\% \\
\addlinespace[2pt]
\fusionss{} (SAX+SFA) & 83.46 $\pm$ 12.30 & -8.01 & +2.48 & 12/33/67 & 26.7\% \\
\bottomrule
\end{tabular}
\end{table}

While both two-way fusions are competitive overall, their behaviour diverges sharply across regimes. Table~\ref{tab:overall-clusters-all} reports regime-level results using robust and Bayesian statistics. In frequency-structured regimes (C1 HighImb and C4 HighFlCx), \fusionsfr{} performs strongly, confirming that SFA contributes substantial gains when spectral cues are informative. In contrast, \fusionsr{} underperforms in C4, highlighting the limited value of time-domain discretisation when frequency complexity dominates.

In regimes characterised by short or locally irregular signals, the pattern reverses. In C6 (ShortBase), \fusionsr{} is the significant two-way winner (HL +0.15 pp, Holm $p=0.0063$), closely matching F3, while \fusionsfr{} adds little value. A similar pattern appears in C2 (LongFSTCx), where \fusionsr{} achieves the largest two-way gain (+0.49 pp), nearly matching F3 (+0.42 pp). These regimes show that adding SAX to ROCKET is often sufficient, and that introducing SFA can be redundant or destabilising.

\begin{table}[t]
\centering
\caption{Overall and per-regime summary vs.\ \textbf{R} (ROCKET). All numbers rounded to 2 decimals.
\hlwin{Green}: Strict Winner (HL$>0$, $p<.05$, ROPE$\ge0.5$).
$\blacktriangle$: Competitive Leader (Best HL or Significant, but ROPE$<0.5$).}
\label{tab:overall-clusters-all}
\scriptsize
\setlength{\tabcolsep}{2pt}
\renewcommand{\arraystretch}{1.1}
\begin{tabular}{l l r c c c c}
\toprule
Regime & Model vs R & $N$ & HL $\Delta$pp [95\% CI] & $p_{\text{Holm}}$ & $P(d{>}0)$ & ROPE \\
\midrule
\multirow{3}{*}{Overall} 
 & \hlwin{F3} & \multirow{3}{*}{113} & \hlwin{+0.43 [0.31, 0.57]} & 0.00 & 0.87 & 0.55 \\
 & \fusionsr{} & & +0.11 [0.04, 0.18] & 0.00 & 0.73 & 0.31 \\
 & \fusionsfr{} & & +0.20 [0.07, 0.35] & 0.00 & 0.77 & 0.38 \\
\midrule
\multirow{3}{*}{C1 HighImb} 
 & \hlwin{F3} & \multirow{3}{*}{38} & \hlwin{+0.51 [0.36, 0.78]} & 0.00 & 0.92 & 0.62 \\
 & \fusionsr{} & & +0.11 [0.00, 0.29] & 0.11 & 0.74 & 0.27 \\
 & \fusionsfr{} & & +0.30 [0.10, 0.50] & 0.00 & 0.95 & 0.39 \\
\midrule
\multirow{3}{*}{C2 LongFSTCx} 
 & F3 & \multirow{3}{*}{11} & +0.42 [0.07, 1.53] & 0.19 & 0.77 & 0.44 \\
 & \textbf{\fusionsr{}} $\blacktriangle$ & & \textbf{+0.49} [0.11, 1.04] & 0.11 & 0.71 & 0.44 \\
 & \fusionsfr{} & & +0.02 [-0.77, 0.97] & 1.00 & 0.50 & 0.28 \\
\midrule
\multirow{3}{*}{C3 SmoothSep} 
 & \hlwin{F3} & \multirow{3}{*}{24} & \hlwin{+0.58 [0.36, 0.86]} & 0.00 & 0.98 & 0.77 \\
 & \fusionsr{} & & +0.09 [0.01, 0.16] & 0.08 & 0.71 & 0.26 \\
 & \fusionsfr{} & & +0.30 [0.06, 0.45] & 0.03 & 0.79 & 0.49 \\
\midrule
\multirow{3}{*}{C4 HighFlCx} 
 & \textbf{F3} $\blacktriangle$ & \multirow{3}{*}{7} & \textbf{+1.09} [-0.63, 2.87] & 0.44 & 0.81 & 0.41 \\
 & \fusionsr{} & & -0.32 [-0.93, 0.25] & 0.75 & 0.56 & 0.18 \\
 & \fusionsfr{} & & +0.80 [-1.17, 2.69] & 0.59 & 0.69 & 0.41 \\
\midrule
\multirow{3}{*}{C5 HighCompOut} 
 & F3 & \multirow{3}{*}{5} & -0.72 [-4.28, 0.05] & 0.44 & 0.30 & 0.08 \\
 & \fusionsr{} & & -0.22 [-1.50, 0.17] & 0.75 & 0.30 & 0.23 \\
 & \fusionsfr{} & & -0.89 [-8.24, -0.22] & 0.38 & 0.10 & 0.08 \\
\midrule
\multirow{3}{*}{C6 ShortBase} 
 & F3 & \multirow{3}{*}{28} & +0.16 [0.01, 0.43] & 0.06 & 0.78 & 0.39 \\
 & \textbf{\fusionsr{}} $\blacktriangle$ & & \textbf{+0.15} [0.04, 0.37] & 0.01 & 0.87 & 0.42 \\
 & \fusionsfr{} & & +0.14 [0.00, 0.37] & 0.08 & 0.76 & 0.36 \\
\bottomrule
\end{tabular}
\end{table}

To further clarify these regime-dependent behaviours, we compute SHAP attributions for the two-way fusion variants. For \fusionsfr{}, spectral measures such as \texttt{spectral\_entropy} and \texttt{global\_kl\_psd}, together with \texttt{ts\_length}, are strongly positively associated with gains, while \texttt{permutation\_entropy} and \texttt{turning\_points} emerges as the dominant failure mode. In contrast, \fusionsr{} shows reduced sensitivity to both of these and emphasises time and frequency domain separability measures, indicating that SAX moderates the negative effects of local disturbances observed in the frequency-only fusion. These patterns align with the relative performance of the two variants across regimes, particularly their contrasting behaviour in C2, C5, and C6 (Table~\ref{tab:overall-clusters-all}). SHAP visualisations for the ablations are provided in Appendix D.4, Figures~\ref{fig:shap2_sr} and~\ref{fig:shap2_sfr}. It is worth mentioning that this behaviour is also reflected in the heatmap (Fig.~\ref{fig:heatmap}) where C5(characterised by high permutation entropy and turning\_points measures and lowest separability measures), where solo SFA performance is strikingly worse ($\approx 61.9\%$) and surprisingly SAX is noticeably better ($\approx 71.8\%$).

Taken together, the ablation results show that ROCKET is the essential backbone, while SFA and SAX contribute complementary but asymmetric value. SFA delivers strong gains when frequency structure is rich, but is vulnerable to local irregularity; SAX provides weaker but stabilising contributions. F3 combines these complementary effects through adaptive gating. When frequency structure is informative, it activates SFA and achieves gains similar to those observed in \fusionsfr{}. When local irregularity increases, SAX mitigates the resulting brittleness, preventing the sharp performance drops seen in the SFA-only ablation. This interaction explains why F3 can outperform both two-way variants in certain regimes, despite the weak standalone performance of SAX and SFA.


\subsection{Case Studies: Sample-Level Mechanisms}
\label{sec:case}

The preceding analyses identify \emph{when} fusion is likely to help (via meta-features and regimes) and \emph{why} its components interact as they do (via ablations). We now examine \emph{how} these effects manifest at the sample level by inspecting representative datasets spanning all regimes. Dataset selection prioritises regime coverage and interpretability rather than optimising performance; however, in several cases, the corrective behaviour is particularly visible due to substantial relative error reductions (e.g., \emph{Rock}, \emph{HouseTwenty}, and \emph{PigArtPressure}). Our focus is on three questions: which samples are corrected by fusion, whether gating behaviour differs systematically on those corrected samples, and how specific confusion patterns change. A comprehensive analysis of gate value relationships with fusion benefit across all clusters is provided in Appendix D.4, Figure~\ref{fig:gate-cluster-grid}.

Table~\ref{tab:case_studies} summarises the key observations. For each dataset, we report the accuracy gain of F3 over ROCKET, the number of \emph{rescued} and \emph{hurt} samples, and the change in average gating weights on rescued samples relative to samples correctly classified by both models. Across all case studies, fusion gains arise primarily from rescued samples, while the number of hurt samples remains small. This indicates that improvements are driven by targeted corrections rather than broad shifts in decision boundaries, consistent with the global patterns reported earlier.

\begin{table}[t]
\centering
\caption{Case-study summary by regime. $\Delta$pp is F3$-$ROCKET accuracy in percentage points. 
Res/Hurt are sample counts. ``Gate shift'' is the change in mean gating weights on \emph{rescued} vs.\ \emph{both\_correct} samples (SAX, SFA, R).}
\label{tab:case_studies}
\footnotesize
\begin{tabular}{lrrrr}
\toprule
Dataset (Cluster) & $N$ & $\Delta$pp & Res/Hurt & Gate shift (SX, SF, R) \\
\midrule
\multicolumn{5}{l}{\textbf{C1: HighImb}}\\
Worms (C1)                    & 256 & 3.12 & 19/11 & (+0.08,\ +0.09,\ -0.18) \\
OliveOil (C1)                 &  58 & 1.72 &  1/0  & (-0.08,\ -0.12,\ +0.20) \\
\addlinespace[2pt]
\multicolumn{5}{l}{\textbf{C2: LongFSTCx}}\\
Rock (C2)                     &  68 & 8.82 &  7/1  & (+0.00,\ +0.10,\ -0.10) \\
PigArtPressure (C2)           & 310 & 1.94 &  8/2  & (+0.00,\ +0.16,\ -0.16) \\
\addlinespace[2pt]
\multicolumn{5}{l}{\textbf{C3: SmoothSep}}\\
Beef (C3)                     &  58 & 3.45 &  5/3  & (+0.00,\ +0.31,\ -0.31) \\
Car (C3)                      & 118 & 1.69 &  3/1  & (+0.00,\ +0.01,\ -0.01) \\
\addlinespace[2pt]
\multicolumn{5}{l}{\textbf{C4: HighFlCx}}\\
RefrigerationDevices (C4)     & 748 & 4.68 & 83/48 & (+0.00,\ +0.08,\ -0.08) \\
\addlinespace[2pt]
\multicolumn{5}{l}{\textbf{C5: HighCompOut}}\\
SemgHandGenderCh2 (C5)        & 898 & 0.11 & 11/10 & (+0.00,\ +0.05,\ -0.04) \\
\addlinespace[2pt]
\multicolumn{5}{l}{\textbf{C6: ShortBase}}\\
HouseTwenty (C6)              & 157 & 1.91 &  3/0  & (-0.03,\ +0.21,\ -0.18) \\
ACSF1 (C6)                    & 198 & 2.02 &  6/2  & (+0.00,\ +0.08,\ -0.08) \\
\bottomrule
\end{tabular}
\end{table}

In representative datasets from frequency-structured regimes (C2 and C4), such as \emph{Rock} and \emph{RefrigerationDevices}, large accuracy gains coincide with clear increases in SFA gate weight on rescued samples. These shifts occur precisely where ROCKET’s errors are corrected, illustrating how frequency-domain cues resolve confusions that persist under random convolutional features alone. This behaviour is consistent with the regime-level and SHAP-based analyses, which associate fusion gains with longer series and richer spectral structure.

In representative datasets from smoothly separable regimes (C3), including \emph{Beef} and \emph{Car}, gains are more modest but still systematic. Fusion corrects a small number of residual errors, with modest increases in SFA weight on rescued samples and minimal disruption to already correct predictions. These cases illustrate how fusion can refine ROCKET’s decisions even when baseline performance is strong.

In representative short or near-ceiling datasets from C6, such as \emph{HouseTwenty} and \emph{ACSF1}, improvements are small but reliable, and hurt rates remain negligible. These examples are consistent with the ablation results, indicating that adding SAX to ROCKET provides a conservative refinement in time-domain-dominated settings.

Finally, a representative dataset from the outlier-heavy regime (C5), \emph{SemgHandGenderCh2}, exhibits neither large gains nor systematic gate shifts. Rescued and hurt samples occur in similar numbers, and gating behaviour shows little change, illustrating a setting in which fusion provides limited benefit.

Overall, the case studies provide concrete, sample-level illustrations of the mechanisms inferred from meta-feature attribution and ablation analyses. They do not serve to generalise beyond the regimes already established, but to make the corrective behaviour of fusion interpretable and verifiable at the level of individual predictions. A detailed case study of the \emph{Rock} dataset, including confusion matrix analysis and gate behaviour visualisation, is provided in Appendix D.5, Figure~\ref{fig:rock-case}.

\subsection{Practical Recommendations}

The analyses suggest that fusion should be applied selectively rather than universally. A simple regime-aware strategy provides reliable guidance:

\begin{itemize}\itemsep0.2em
  \item \textbf{Use F3 (SAX+SFA+ROCKET)} when datasets exhibit heterogeneous structure across time and frequency domains. This includes regimes such as \textbf{HighImb}, \textbf{SmoothSep}, and \textbf{HighFlCx}, where complementary cues are consistently available and three-way fusion yields the most reliable improvements.

  \item \textbf{Use \fusionsr{} (SAX+ROCKET)} when shape-based time-domain cues dominate or when computational efficiency is a priority. This setting applies to \textbf{LongFSTCx} and \textbf{ShortBase}, where two-way fusion achieves performance comparable to F3 at lower cost.

  \item \textbf{Use ROCKET alone} in highly irregular, outlier-heavy settings (\textbf{HighCompOut}), or when baseline accuracy is already very high ($>95\%$), where fusion offers limited or inconsistent benefit.
\end{itemize}

These recommendations emphasise matching model complexity to data structure rather than treating fusion as a universal upgrade.

\section{Conclusion}
\label{sec:conclusion}

We revisited univariate time series classification from a \emph{regime-aware} perspective, asking not whether fusion improves performance on average, but \emph{when and why} it does so. Using meta-features to characterise datasets, we identified six interpretable regimes that explain much of the observed heterogeneity in ROCKET’s performance and in the effectiveness of representation fusion.

Across 113 UCR datasets, three-way fusion (F3) delivers small but consistent accuracy improvements over ROCKET, supported by robust paired statistics and Bayesian evidence. Importantly, these gains are not uniform. They concentrate in regimes where signals are long, spectrally diverse, or exhibit structured variability, and diminish—or reverse—in settings dominated by local irregularity or extreme noise.

Global SHAP attribution clarifies which dataset properties predict fusion gains, while ablation studies isolate the asymmetric roles of the constituent representations. Frequency-domain features (SFA) provide strong but brittle gains when spectral structure is informative, whereas time-domain symbolic features (SAX) offer weaker but stabilising contributions. F3 succeeds by adaptively balancing these effects around a strong convolutional backbone, rather than by replacing it.

Sample-level case studies further confirm this mechanism: fusion improves performance primarily by rescuing specific errors, with gate weights shifting toward frequency-based representations exactly where corrections occur. Conversely, in outlier-heavy regimes, fusion fails in an interpretable way, reinforcing that fusion should be applied selectively rather than indiscriminately.

Overall, the central message is pragmatic: \emph{small, consistent, and explainable gains are preferable to sporadic large wins}. By tying meta-features to regimes, validating improvements with robust statistics, and exposing mechanisms through attribution, ablations, and sample-level analysis, regime-aware fusion offers a dependable extension to strong baselines like ROCKET—precisely where the data support it.

\paragraph{Limitations and future work.}
Our conclusions are affected by small sample sizes in some regimes (notably C4 and C5) and are restricted to univariate datasets from the UCR archive. The fusion architecture is intentionally simple to preserve interpretability. Future work includes automatic regime prediction for zero-shot model selection, extensions to multivariate and irregular time series, budget-aware or instance-level gating strategies, and exploration of additional representations and fusion mechanisms.

\section*{Acknowledgments}
We thank the University of Bristol Advanced Computing Research Centre for providing HPC resources that made the extensive hyperparameter grid searches feasible across all 113 datasets. We also acknowledge the UCR Time Series Classification Archive \cite{Dau2019UCR} for providing the benchmark datasets used in this study.

\appendix
\clearpage
\section*{Appendix}
\subsection*{A. Accuracy Variability Table (Mean$\pm$SD)}
\label{app:acc-sd}

\begingroup
\scriptsize
\setlength{\tabcolsep}{5pt}  

\begin{longtable}{l r l c r r}
\caption{Accuracy by cluster and model (UCR subset, per-dataset means aggregated within clusters).
Acc(\%) shown as mean $\pm$ SD across datasets in the cluster.
$\Delta$pp = (Model $-$ ROCKET) in percentage points;
$\Delta$SD = SD$_{\text{Model}} -$ SD$_{\text{ROCKET}}$ (pp; negative = more stable).}
\label{tab:appendix_all_models_by_cluster}\\
\toprule
Cluster (regime) & N & Model & Acc (\%) & $\Delta$pp & $\Delta$SD \\
\midrule
\endfirsthead
\toprule
Cluster (regime) & N & Model & Acc (\%) & $\Delta$pp & $\Delta$SD \\
\midrule
\endhead
\midrule
\multicolumn{6}{r}{\emph{Table continues on next page}}\\
\midrule
\endfoot
\bottomrule
\endlastfoot
C1 (HighImb) & 38 & ROCKET        & 91.84 $\pm$ 8.47 & +0.00 & +0.00 \\
             &    & F3            & 92.46 $\pm$ 8.04 & +0.62 & -0.43 \\
             &    & \fusionsfr{}        & 92.21 $\pm$ 8.21 & +0.37 & -0.26 \\
             &    & \fusionsr{}         & 91.95 $\pm$ 8.39 & +0.11 & -0.08 \\
             &    & SFA           & 88.21 $\pm$ 10.25 & -3.63 & +1.78 \\
             &    & SAX           & 76.65 $\pm$ 13.30 & -15.19 & +4.83 \\
C2 (LongFSTCx) & 11 & ROCKET       & 83.78 $\pm$ 11.46 & +0.00 & +0.00 \\
               &    & F3           & 84.93 $\pm$ 12.04 & +1.15 & +0.58 \\
               &    & \fusionsfr{}       & 84.21 $\pm$ 12.15 & +0.43 & +0.69 \\
               &    & \fusionsr{}        & 84.39 $\pm$ 11.38 & +0.61 & -0.08 \\
               &    & SFA          & 64.23 $\pm$ 27.49 & -19.55 & +16.03 \\
               &    & SAX          & 59.14 $\pm$ 25.42 & -24.64 & +13.96 \\
C3 (SmoothSep) & 24 & ROCKET        & 92.56 $\pm$ 7.99 & +0.00 & +0.00 \\
               &    & F3            & 93.24 $\pm$ 7.57 & +0.68 & -0.42 \\
               &    & \fusionsfr{}       & 92.81 $\pm$ 8.09 & +0.25 & +0.10 \\
               &    & \fusionsr{}        & 92.68 $\pm$ 7.97 & +0.12 & -0.02 \\
               &    & SFA           & 78.80 $\pm$ 19.73 & -13.76 & +11.74 \\
               &    & SAX           & 68.23 $\pm$ 18.65 & -24.33 & +10.66 \\
C4 (HighFlCx)  &  7 & ROCKET        & 79.34 $\pm$ 16.88 & +0.00 & +0.00 \\
               &    & F3            & 80.43 $\pm$ 16.40 & +1.09 & -0.48 \\
               &    & \fusionsfr{}        & 80.14 $\pm$ 16.19 & +0.80 & -0.69 \\
               &    & \fusionsr{}         & 79.03 $\pm$ 16.82 & -0.31 & -0.06 \\
               &    & SFA           & 71.17 $\pm$ 16.16 & -8.17 & -0.72 \\
               &    & SAX           & 59.10 $\pm$ 16.13 & -20.24 & -0.75 \\
C5 (HighCompOut) & 5 & ROCKET       & 91.64 $\pm$ 8.12 & +0.00 & +0.00 \\
                 &   & F3           & 89.91 $\pm$ 9.79 & -1.73 & +1.67 \\
                 &   & \fusionsfr{}       & 88.08 $\pm$ 12.36 & -3.56 & +4.24 \\
                 &   & \fusionsr{}        & 91.09 $\pm$ 8.84 & -0.55 & +0.72 \\
                 &   & SFA          & 61.96 $\pm$ 31.14 & -29.68 & +23.02 \\
                 &   & SAX          & 71.84 $\pm$ 13.86 & -19.80 & +5.74 \\
C6 (ShortBase)  & 28 & ROCKET        & 96.05 $\pm$ 6.33 & +0.00 & +0.00 \\
               &    & F3            & 96.26 $\pm$ 6.31 & +0.21 & -0.02 \\
               &    & \fusionsfr{}        & 96.24 $\pm$ 6.27 & +0.19 & -0.06 \\
               &    & \fusionsr{}         & 96.27 $\pm$ 6.16 & +0.22 & -0.17 \\
               &    & SFA           & 86.02 $\pm$ 19.08 & -10.03 & +12.75 \\
               &    & SAX           & 75.04 $\pm$ 22.22 & -21.01 & +15.89 \\
All datasets   & 113 & ROCKET        & 91.47 $\pm$ 9.82 & +0.00 & +0.00 \\
               &     & F3            & 91.98 $\pm$ 9.59 & +0.51 & -0.23 \\
               &     & \fusionsfr{}        & 91.63 $\pm$ 9.89 & +0.16 & +0.07 \\
               &     & \fusionsr{}         & 91.60 $\pm$ 9.78 & +0.13 & -0.04 \\
               &     & SFA           & 81.12 $\pm$ 19.83 & -10.35 & +10.01 \\
               &     & SAX           & 71.46 $\pm$ 19.14 & -20.01 & +9.32 \\
\end{longtable}
\endgroup

\clearpage
\subsection*{B. Per-Dataset Accuracy Details}
\label{app:dataset-accuracies}
\begingroup
\scriptsize
\setlength{\tabcolsep}{2pt}

\begin{longtable}{l r l r r r r r r r}
\caption{Per-dataset accuracies and cluster assignments. Accuracies are rounded to three decimal digits.}
\label{tab:dataset-accuracies-by-cluster}\\
\toprule
Dataset & Cluster & ClusterName & SAX & SFA & ROCKET & \fusionsfr{} & \fusionsr{} & \fusionss{} & F3 \\
\midrule
\endfirsthead

\toprule
Dataset & Cluster & ClusterName & SAX & SFA & ROCKET & \fusionsfr{} & \fusionsr{} & \fusionss{} & F3 \\
\midrule
\endhead

\midrule
\multicolumn{10}{r}{{Continued on next page}} \\
\midrule
\endfoot

\bottomrule
\endlastfoot

ACSF1 & 6 & ShortBase & 0.541 & 0.823 & 0.868 & 0.874 & 0.884 & 0.848 & 0.889 \\
Adiac & 3 & SmoothSep & 0.195 & 0.754 & 0.855 & 0.852 & 0.858 & 0.752 & 0.861 \\
ArrowHead & 1 & HighImb & 0.770 & 0.871 & 0.952 & 0.957 & 0.952 & 0.876 & 0.962 \\
BME & 6 & ShortBase & 0.854 & 0.943 & 1.000 & 1.000 & 1.000 & 0.961 & 1.000 \\
Beef & 3 & SmoothSep & 0.518 & 0.806 & 0.777 & 0.758 & 0.776 & 0.811 & 0.808 \\
BeetleFly & 1 & HighImb & 0.871 & 0.950 & 0.921 & 0.921 & 0.896 & 0.975 & 0.921 \\
BirdChicken & 1 & HighImb & 0.900 & 1.000 & 1.000 & 1.000 & 1.000 & 1.000 & 1.000 \\
CBF & 6 & ShortBase & 0.970 & 0.994 & 1.000 & 1.000 & 1.000 & 0.997 & 1.000 \\
Car & 3 & SmoothSep & 0.638 & 0.814 & 0.924 & 0.932 & 0.932 & 0.830 & 0.941 \\
Chinatown & 1 & HighImb & 0.717 & 0.964 & 0.986 & 0.986 & 0.989 & 0.967 & 0.989 \\
ChlorineConcentration & 6 & ShortBase & 0.573 & 0.993 & 0.995 & 0.993 & 0.991 & 0.992 & 0.993 \\
CinCECGTorso & 2 & LongFSTCx & 0.898 & 0.999 & 0.999 & 1.000 & 0.997 & 1.000 & 1.000 \\
Coffee & 1 & HighImb & 0.871 & 1.000 & 1.000 & 1.000 & 1.000 & 1.000 & 1.000 \\
Computers & 4 & HighFlCx & 0.689 & 0.829 & 0.906 & 0.874 & 0.894 & 0.845 & 0.882 \\
CricketX & 6 & ShortBase & 0.515 & 0.510 & 0.855 & 0.857 & 0.861 & 0.512 & 0.859 \\
CricketY & 6 & ShortBase & 0.496 & 0.445 & 0.862 & 0.866 & 0.863 & 0.460 & 0.865 \\
CricketZ & 6 & ShortBase & 0.523 & 0.481 & 0.852 & 0.861 & 0.853 & 0.488 & 0.862 \\
Crop & 6 & ShortBase & 0.042 & 0.675 & 0.775 & 0.767 & 0.777 & 0.677 & 0.772 \\
DiatomSizeReduction & 3 & SmoothSep & 0.931 & 0.997 & 1.000 & 1.000 & 1.000 & 0.997 & 1.000 \\
DistalPhalanxOutlineAgeGroup & 1 & HighImb & 0.730 & 0.827 & 0.849 & 0.858 & 0.849 & 0.830 & 0.858 \\
DistalPhalanxOutlineCorrect & 1 & HighImb & 0.684 & 0.828 & 0.856 & 0.865 & 0.864 & 0.835 & 0.863 \\
DistalPhalanxTW & 1 & HighImb & 0.672 & 0.780 & 0.804 & 0.810 & 0.810 & 0.784 & 0.814 \\
ECG200 & 1 & HighImb & 0.834 & 0.869 & 0.944 & 0.944 & 0.939 & 0.884 & 0.955 \\
ECG5000 & 1 & HighImb & 0.939 & 0.943 & 0.958 & 0.958 & 0.958 & 0.943 & 0.958 \\
ECGFiveDays & 6 & ShortBase & 0.896 & 0.999 & 1.000 & 1.000 & 1.000 & 0.999 & 1.000 \\
EOGHorizontalSignal & 2 & LongFSTCx & 0.554 & 0.302 & 0.841 & 0.845 & 0.846 & 0.514 & 0.846 \\
EOGVerticalSignal & 2 & LongFSTCx & 0.493 & 0.201 & 0.798 & 0.791 & 0.809 & 0.442 & 0.805 \\
Earthquakes & 5 & HighCompOut & 0.800 & 0.802 & 0.808 & 0.804 & 0.804 & 0.802 & 0.806 \\
ElectricDevices & 4 & HighFlCx & 0.625 & 0.860 & 0.901 & 0.900 & 0.904 & 0.869 & 0.903 \\
EthanolLevel & 2 & LongFSTCx & 0.392 & 0.494 & 0.801 & 0.780 & 0.797 & 0.487 & 0.797 \\
FaceAll & 6 & ShortBase & 0.679 & 0.958 & 0.992 & 0.993 & 0.993 & 0.957 & 0.994 \\
FaceFour & 6 & ShortBase & 0.945 & 0.982 & 1.000 & 1.000 & 1.000 & 0.991 & 1.000 \\
FacesUCR & 6 & ShortBase & 0.690 & 0.956 & 0.993 & 0.993 & 0.993 & 0.952 & 0.993 \\
FiftyWords & 3 & SmoothSep & 0.456 & 0.434 & 0.843 & 0.847 & 0.843 & 0.485 & 0.858 \\
Fish & 3 & SmoothSep & 0.557 & 0.934 & 0.960 & 0.965 & 0.963 & 0.940 & 0.965 \\
FordA & 1 & HighImb & 0.801 & 0.918 & 0.943 & 0.951 & 0.943 & 0.928 & 0.952 \\
FordB & 1 & HighImb & 0.757 & 0.903 & 0.927 & 0.934 & 0.928 & 0.908 & 0.935 \\
FreezerRegularTrain & 6 & ShortBase & 0.852 & 0.976 & 1.000 & 1.000 & 1.000 & 0.979 & 1.000 \\
FreezerSmallTrain & 6 & ShortBase & 0.850 & 0.977 & 1.000 & 1.000 & 1.000 & 0.982 & 1.000 \\
Fungi & 3 & SmoothSep & 0.679 & 0.955 & 1.000 & 1.000 & 1.000 & 0.960 & 1.000 \\
GunPoint & 1 & HighImb & 0.944 & 1.000 & 1.000 & 1.000 & 1.000 & 1.000 & 1.000 \\
GunPointAgeSpan & 1 & HighImb & 0.949 & 0.989 & 0.996 & 0.998 & 0.996 & 0.993 & 0.998 \\
GunPointMaleVersusFemale & 1 & HighImb & 0.964 & 0.993 & 0.998 & 0.998 & 0.998 & 0.996 & 1.000 \\
GunPointOldVersusYoung & 1 & HighImb & 0.920 & 0.991 & 0.996 & 0.998 & 0.998 & 0.993 & 1.000 \\
Ham & 1 & HighImb & 0.778 & 0.788 & 0.863 & 0.887 & 0.878 & 0.807 & 0.882 \\
HandOutlines & 2 & LongFSTCx & 0.828 & 0.791 & 0.950 & 0.950 & 0.954 & 0.835 & 0.952 \\
Haptics & 2 & LongFSTCx & 0.471 & 0.490 & 0.675 & 0.679 & 0.681 & 0.521 & 0.683 \\
Herring & 1 & HighImb & 0.675 & 0.706 & 0.755 & 0.763 & 0.755 & 0.714 & 0.763 \\
HouseTwenty & 6 & ShortBase & 0.949 & 0.987 & 0.975 & 0.988 & 0.981 & 0.988 & 0.994 \\
InlineSkate & 2 & LongFSTCx & 0.343 & 0.463 & 0.767 & 0.752 & 0.769 & 0.472 & 0.770 \\
InsectEPGRegularTrain & 6 & ShortBase & 0.880 & 0.977 & 0.997 & 0.997 & 1.000 & 0.984 & 1.000 \\
InsectEPGSmallTrain & 6 & ShortBase & 0.879 & 0.977 & 1.000 & 0.996 & 1.000 & 0.981 & 1.000 \\
InsectWingbeatSound & 3 & SmoothSep & 0.583 & 0.471 & 0.724 & 0.725 & 0.726 & 0.580 & 0.729 \\
ItalyPowerDemand & 1 & HighImb & 0.502 & 0.969 & 0.978 & 0.978 & 0.980 & 0.970 & 0.979 \\
LargeKitchenAppliances & 4 & HighFlCx & 0.710 & 0.659 & 0.938 & 0.940 & 0.939 & nan & 0.940 \\
Lightning2 & 6 & ShortBase & 0.840 & 0.698 & 0.908 & 0.916 & 0.916 & 0.714 & 0.874 \\
Lightning7 & 6 & ShortBase & 0.611 & 0.447 & 0.893 & 0.901 & 0.901 & 0.461 & 0.908 \\
Mallat & 3 & SmoothSep & 0.968 & 1.000 & 0.997 & 0.998 & 0.998 & 1.000 & 0.998 \\
Meat & 3 & SmoothSep & 0.678 & 0.957 & 1.000 & 1.000 & 1.000 & 0.966 & 1.000 \\
MedicalImages & 1 & HighImb & 0.602 & 0.648 & 0.878 & 0.881 & 0.882 & 0.644 & 0.887 \\
MiddlePhalanxOutlineAgeGroup & 1 & HighImb & 0.708 & 0.763 & 0.770 & 0.777 & 0.775 & 0.764 & 0.777 \\
MiddlePhalanxOutlineCorrect & 1 & HighImb & 0.661 & 0.811 & 0.867 & 0.870 & 0.870 & 0.810 & 0.874 \\
MiddlePhalanxTW & 1 & HighImb & 0.579 & 0.648 & 0.653 & 0.664 & 0.661 & 0.650 & 0.675 \\
MixedShapesRegularTrain & 3 & SmoothSep & 0.858 & 0.966 & 0.978 & 0.987 & 0.980 & 0.972 & 0.989 \\
MixedShapesSmallTrain & 3 & SmoothSep & 0.858 & 0.962 & 0.980 & 0.985 & 0.980 & 0.967 & 0.987 \\
MoteStrain & 1 & HighImb & 0.882 & 0.979 & 0.992 & 0.992 & 0.992 & 0.983 & 0.993 \\
NonInvasiveFetalECGThorax1 & 3 & SmoothSep & 0.596 & 0.557 & 0.969 & 0.971 & 0.970 & 0.580 & 0.972 \\
NonInvasiveFetalECGThorax2 & 3 & SmoothSep & 0.664 & 0.645 & 0.968 & 0.970 & 0.971 & 0.660 & 0.971 \\
OSULeaf & 3 & SmoothSep & 0.775 & 0.982 & 0.957 & 0.966 & 0.957 & 0.980 & 0.968 \\
OliveOil & 1 & HighImb & 0.503 & 0.747 & 0.932 & 0.948 & 0.948 & 0.692 & 0.948 \\
PhalangesOutlinesCorrect & 1 & HighImb & 0.663 & 0.805 & 0.863 & 0.855 & 0.868 & 0.808 & 0.859 \\
Phoneme & 4 & HighFlCx & 0.277 & 0.419 & 0.463 & 0.482 & 0.465 & 0.435 & 0.475 \\
PigAirwayPressure & 2 & LongFSTCx & 0.123 & 0.613 & 0.645 & 0.645 & 0.655 & 0.606 & 0.645 \\
PigArtPressure & 2 & LongFSTCx & 0.900 & 0.990 & 0.958 & 0.971 & 0.968 & 0.990 & 0.977 \\
PigCVP & 2 & LongFSTCx & 0.784 & 0.839 & 0.913 & 0.910 & 0.910 & 0.839 & 0.910 \\
Plane & 3 & SmoothSep & 0.971 & 1.000 & 1.000 & 1.000 & 1.000 & 1.000 & 1.000 \\
PowerCons & 6 & ShortBase & 0.925 & 0.905 & 0.986 & 0.989 & 0.989 & 0.916 & 0.992 \\
ProximalPhalanxOutlineAgeGroup & 1 & HighImb & 0.723 & 0.857 & 0.869 & 0.871 & 0.876 & 0.861 & 0.876 \\
ProximalPhalanxOutlineCorrect & 1 & HighImb & 0.705 & 0.859 & 0.904 & 0.916 & 0.913 & 0.858 & 0.916 \\
ProximalPhalanxTW & 1 & HighImb & 0.677 & 0.847 & 0.856 & 0.856 & 0.857 & 0.846 & 0.861 \\
RefrigerationDevices & 4 & HighFlCx & 0.671 & 0.821 & 0.814 & 0.861 & 0.805 & 0.833 & 0.861 \\
Rock & 2 & LongFSTCx & 0.720 & 0.882 & 0.869 & 0.941 & 0.898 & 0.911 & 0.956 \\
ScreenType & 4 & HighFlCx & 0.468 & 0.596 & 0.679 & 0.687 & 0.671 & 0.599 & 0.699 \\
SemgHandGenderCh2 & 5 & HighCompOut & 0.796 & 0.689 & 0.962 & 0.953 & 0.965 & 0.772 & 0.963 \\
SemgHandMovementCh2 & 5 & HighCompOut & 0.485 & 0.253 & 0.854 & 0.703 & 0.829 & 0.464 & 0.783 \\
SemgHandSubjectCh2 & 5 & HighCompOut & 0.698 & 0.354 & 0.958 & 0.944 & 0.957 & 0.684 & 0.943 \\
ShapeletSim & 5 & HighCompOut & 0.813 & 1.000 & 1.000 & 1.000 & 1.000 & 1.000 & 1.000 \\
ShapesAll & 3 & SmoothSep & 0.644 & 0.743 & 0.926 & 0.925 & 0.926 & 0.770 & 0.930 \\
SmallKitchenAppliances & 4 & HighFlCx & 0.697 & 0.797 & 0.852 & 0.866 & 0.854 & 0.817 & 0.870 \\
SmoothSubspace & 6 & ShortBase & 0.336 & 0.624 & 0.980 & 0.983 & 0.983 & 0.651 & 0.986 \\
SonyAIBORobotSurface1 & 6 & ShortBase & 0.876 & 0.990 & 0.997 & 0.998 & 0.998 & 0.990 & 0.998 \\
SonyAIBORobotSurface2 & 6 & ShortBase & 0.806 & 0.987 & 0.998 & 0.999 & 0.999 & 0.988 & 0.999 \\
StarLightCurves & 1 & HighImb & 0.933 & 0.938 & 0.985 & 0.985 & 0.985 & 0.979 & 0.986 \\
Strawberry & 1 & HighImb & 0.784 & 0.978 & 0.985 & 0.987 & 0.984 & 0.979 & 0.985 \\
SwedishLeaf & 3 & SmoothSep & 0.576 & 0.923 & 0.972 & 0.972 & 0.972 & 0.928 & 0.975 \\
Symbols & 3 & SmoothSep & 0.949 & 0.994 & 0.993 & 0.994 & 0.995 & 0.995 & 0.995 \\
SyntheticControl & 6 & ShortBase & 0.834 & 0.926 & 1.000 & 1.000 & 1.000 & 0.933 & 1.000 \\
ToeSegmentation1 & 6 & ShortBase & 0.932 & 0.985 & 0.974 & 0.981 & 0.981 & 0.985 & 0.981 \\
ToeSegmentation2 & 1 & HighImb & 0.897 & 0.945 & 0.975 & 0.976 & 0.976 & 0.951 & 0.982 \\
Trace & 6 & ShortBase & 0.944 & 0.995 & 1.000 & 1.000 & 1.000 & 1.000 & 1.000 \\
TwoLeadECG & 1 & HighImb & 0.684 & 1.000 & 1.000 & 1.000 & 1.000 & 1.000 & 1.000 \\
TwoPatterns & 6 & ShortBase & 0.869 & 0.890 & 1.000 & 1.000 & 1.000 & 0.887 & 1.000 \\
UMD & 6 & ShortBase & 0.905 & 0.983 & 0.994 & 0.994 & 0.994 & 0.989 & 0.994 \\
UWaveGestureLibraryAll & 3 & SmoothSep & 0.785 & 0.737 & 0.983 & 0.983 & 0.983 & 0.854 & 0.984 \\
UWaveGestureLibraryX & 3 & SmoothSep & 0.729 & 0.628 & 0.891 & 0.898 & 0.888 & 0.803 & 0.899 \\
UWaveGestureLibraryY & 3 & SmoothSep & 0.613 & 0.611 & 0.838 & 0.844 & 0.844 & 0.691 & 0.846 \\
UWaveGestureLibraryZ & 3 & SmoothSep & 0.695 & 0.615 & 0.845 & 0.858 & 0.844 & 0.768 & 0.856 \\
Wafer & 1 & HighImb & 0.994 & 1.000 & 1.000 & 1.000 & 1.000 & 1.000 & 1.000 \\
Wine & 1 & HighImb & 0.532 & 0.963 & 1.000 & 1.000 & 1.000 & 0.964 & 1.000 \\
WordSynonyms & 3 & SmoothSep & 0.459 & 0.426 & 0.836 & 0.843 & 0.839 & 0.496 & 0.846 \\
Worms & 1 & HighImb & 0.684 & 0.785 & 0.809 & 0.820 & 0.793 & 0.785 & 0.840 \\
WormsTwoClass & 1 & HighImb & 0.813 & 0.824 & 0.856 & 0.856 & 0.852 & 0.840 & 0.867 \\
Yoga & 1 & HighImb & 0.824 & 0.834 & 0.980 & 0.980 & 0.978 & 0.879 & 0.981 \\

\end{longtable}
\endgroup

\subsection*{C. Meta-Features}

\begin{table}[h!]
\centering
\caption{Mathematical summary of selected dataset-level meta-features.}
\label{tab:meta_features_compact}
\scriptsize
\begin{threeparttable}
\setlength{\tabcolsep}{4pt}
\begin{tabular}{p{2.8cm} p{4.7cm} p{1.5cm} p{4.1cm}}
\hline
\textbf{Meta-Feature} & \textbf{Formula} & \textbf{Level} & \textbf{Higher values indicate} \\ \hline

\texttt{spectral\_entropy}$^{*}$ &
$H=-\sum_k p_k \ln p_k$ &
Global* & More noise-like / complex spectra \\

\texttt{spectral\_entropy\_var} &
$\mathrm{Var}_i\!\left(H^{(i)}\right)$ &
Global & Greater heterogeneity in spectral complexity \\

\texttt{turning\_points}$^{*}$ &
$\tfrac{1}{n}\sum \mathbf{1}\{\Delta x_{t+1}\,\Delta x_t<0\}$ &
Global* & More oscillatory local structure \\

\texttt{turning\_points\_var} &
$\mathrm{Var}_i(\mathrm{TP}^{(i)})$ &
Global & Heterogeneous local volatility \\

\texttt{kurtosis}$^{*}$ &
$\tfrac{\mathbb{E}[(x-\bar x)^4]}{(\mathbb{E}[(x-\bar x)^2])^2}-3$ &
Global* & Heavier tails / spike-like behavior \\

\texttt{autocorr\_lag1}$^{*}$ &
$\rho_1=\mathrm{Corr}(x_t,x_{t+1})$ &
Global* & Stronger short-term temporal dependence \\

\texttt{permutation\_entropy}$^{*}$ &
$-\sum_{\pi} p(\pi)\ln p(\pi)$ &
Global* & Greater ordinal irregularity \\

\texttt{ts\_length} &
$L$ &
Global & Longer temporal context \\

\texttt{kl\_psd}$^{*}$ &
$\mathbb{E}_i\!\left[\mathrm{SKL}(p^{(i)},\bar p)\right]$ &
Global* & Higher global spectral diversity \\

\texttt{dtw\_separability\_time} &
$\tfrac{\mathbb{E}[d_{ij}\mid y_i\neq y_j]}{\mathbb{E}[d_{ij}\mid y_i=y_j]}$ &
Class-based & Easier time-domain class separation \\

\texttt{dtw\_separability\_freq} &
Same ratio applied to PSD sequences &
Class-based & Easier frequency-domain separation \\

\texttt{kruskal\_psd} &
$\mathrm{KW}$ on PSD energy by class &
Class-based & Stronger class spectral differences \\

\texttt{imbalance\_index} &
$\max_c \tfrac{n_c}{N}$ &
Class-based & Stronger class imbalance \\

\hline
\end{tabular}

\begin{tablenotes}[flushleft]
\footnotesize
\item \textit{Notes.}
Features marked with $^{*}$ are computed per time series after z-normalization and
then averaged across samples to obtain a dataset-level summary.
Variance features are computed directly across per-series values and require no
additional aggregation.
Time series length is constant within each dataset and is therefore treated as a
dataset-level attribute.
Power spectral density (PSD) estimates use Welch’s method with
$n_{\text{perseg}}=\min(256,L)$ and natural logarithms.
Class-based DTW separability estimates
$\mathbb{E}[d_{ij}\mid y_i\neq y_j]$ and $\mathbb{E}[d_{ij}\mid y_i=y_j]$
via averages over approximately balanced, dynamically subsampled class pairs,
reducing dominance of large classes and ensuring comparability across datasets.
\end{tablenotes}

\end{threeparttable}
\end{table}

\begin{table}[h!]
\centering
\caption{Selected meta-features used for clustering and their motivation.}
\label{tab:meta_features_justification}
\scriptsize
\setlength{\tabcolsep}{5pt}
\renewcommand{\arraystretch}{1.2}
\begin{tabular}{p{3.0cm} p{9.8cm}}
\hline
\textbf{Feature Group} & \textbf{Included features and rationale} \\
\hline
\textbf{Shape / Statistics} &
\texttt{turning\_points}, \texttt{turning\_points\_var}: capture local oscillatory structure and heterogeneity; 
\texttt{kurtosis}: identifies heavy-tailed or spike-like behavior; 
\texttt{autocorr\_lag1}: measures short-term temporal dependence. \\
\hline
\textbf{Spectral / Entropy} &
\texttt{spectral\_entropy}, \texttt{spectral\_entropy\_var}: quantify frequency-domain complexity and diversity;
\texttt{permutation\_entropy}: measures ordinal unpredictability;
\texttt{kruskal\_psd}, \texttt{global\_kl\_psd}: capture class-wise spectral separation and global spectral diversity. \\
\hline
\textbf{Separability Measures} &
\texttt{dtw\_separability\_time}, \texttt{dtw\_separability\_freq}: quantify class separability in the time and frequency domains using DTW-based distance ratios. \\
\hline
\textbf{Dataset Properties} &
\texttt{ts\_length}: reflects available temporal context;
\texttt{imbalance}: captures class distribution skewness (default accuracy baseline). \\
\hline
\end{tabular}
\end{table}

\clearpage
\FloatBarrier
\subsection*{D. Supplementary Figures}

This section contains supplementary figures referenced in the main text.

\FloatBarrier
\begin{samepage}
\subsubsection{D.1. Cluster Composition}

\vspace{0.2cm}

\centering
\includegraphics[width=\textwidth]{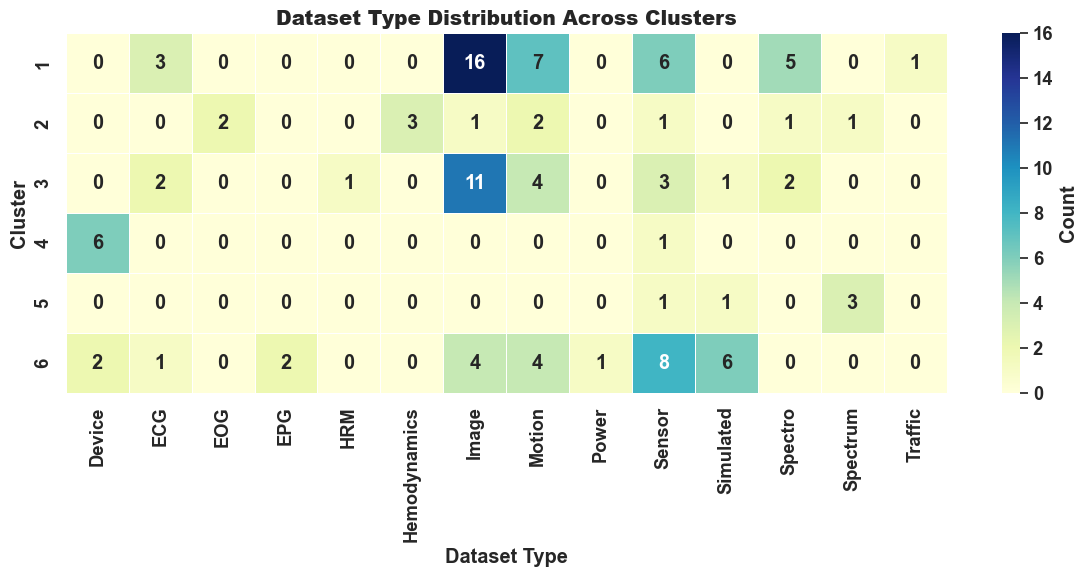}

\captionof{figure}{Distribution of dataset types across clusters for all 113 UCR datasets analyzed. Device datasets concentrate in C4 (often pro-SFA), while image, motion, and sensor datasets dominate C1, C3, and C6 (often pro-ROCKET).}
\label{fig:cdtype_dist}
\end{samepage}

\FloatBarrier
\begin{samepage}
\subsubsection{D.2. 2-D Meta-Feature Projections}

\vspace{0.1cm}

\centering
\includegraphics[width=0.8\textwidth]{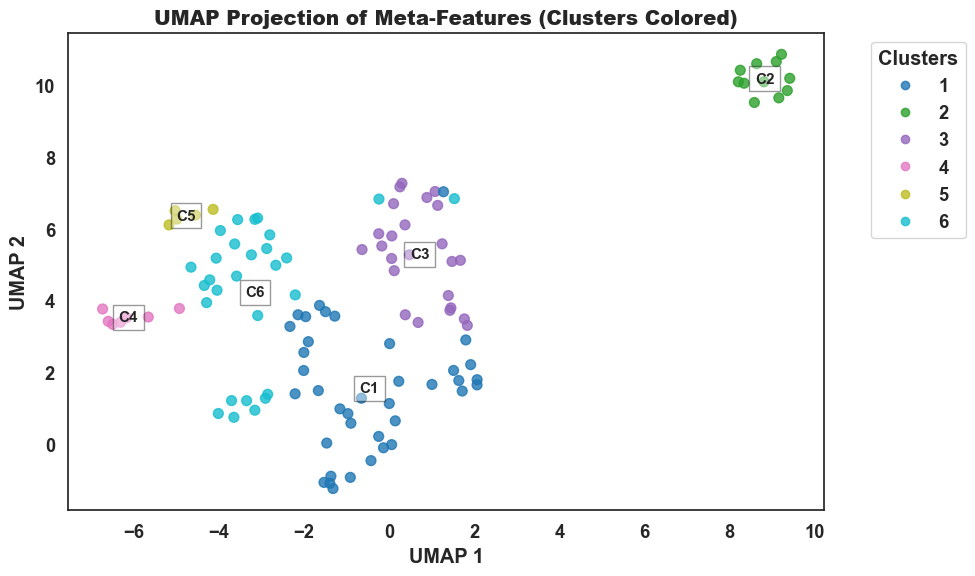}

\vspace{0.1cm}

\centering
\includegraphics[width=0.8\textwidth]{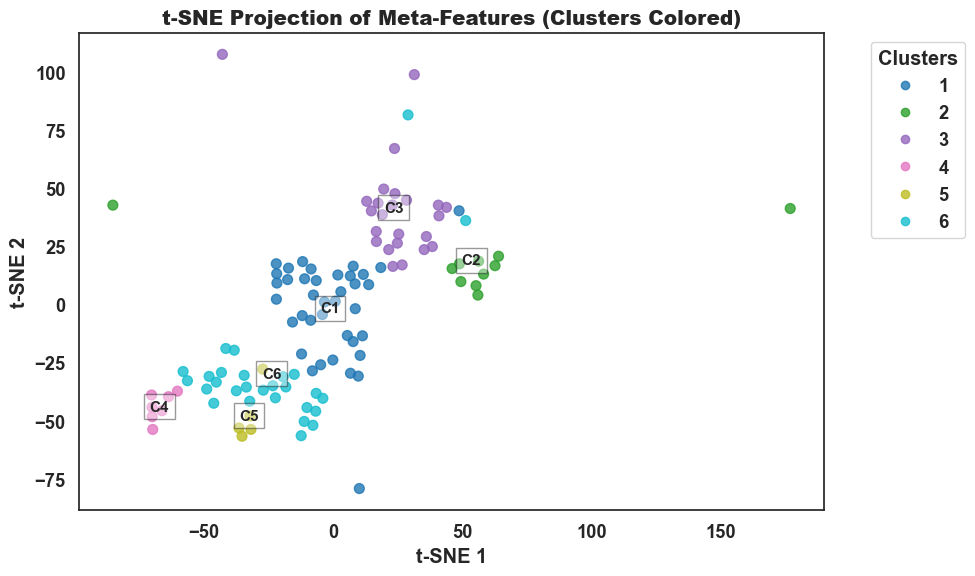}

\captionof{figure}{2-D meta-feature projections using UMAP (top; $n\_neighbors{=}5$, $min\_dist{=}0.36$, Euclidean, seed 18) and t-SNE (bottom; perplexity 11). Labels reflect hierarchical clustering (Ward linkage).}
\label{fig:umap_tsne}
\end{samepage}

\begin{samepage}
\subsubsection{D.3. Gain Distribution}

\vspace{0.5cm}

\centering
\includegraphics[width=\textwidth]{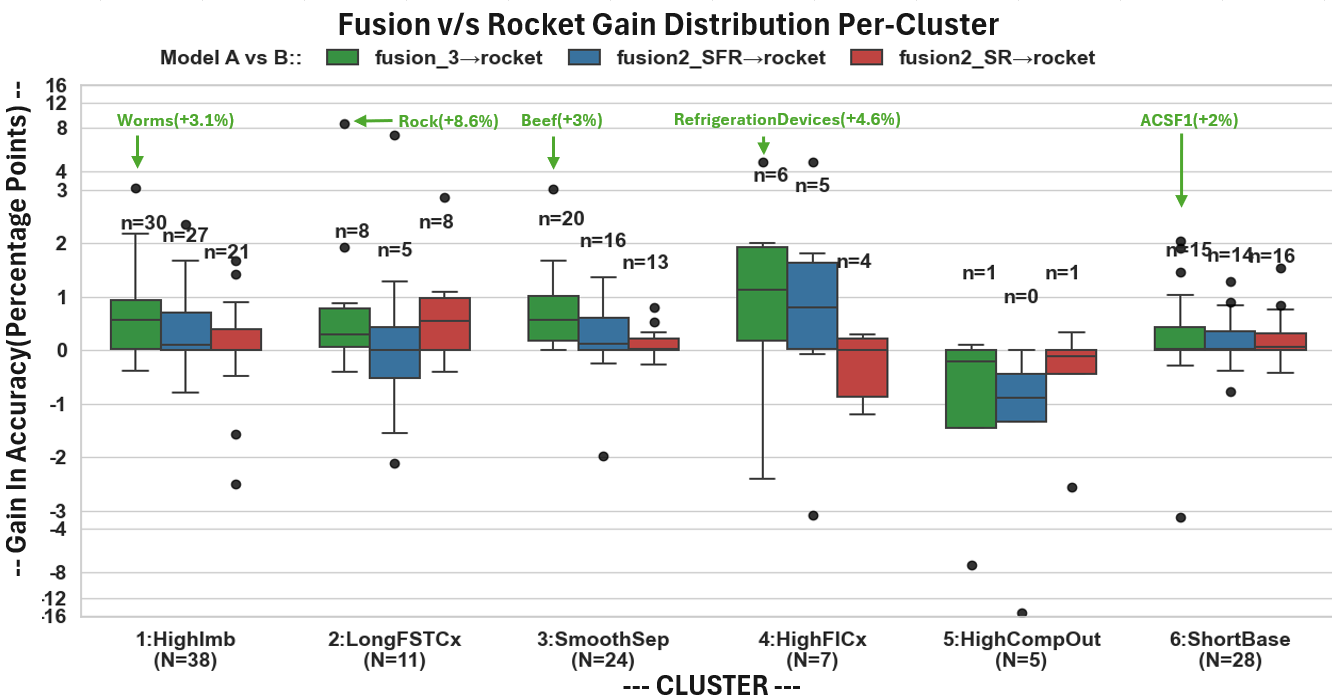}

\captionof{figure}{Distribution of accuracy gains (percentage points) for Fusion models compared to the ROCKET baseline across the six clusters. Y-axis scale in symlog, linear till +/-3. Each boxplot shows the gain/loss for: F3 vs ROCKET (green), \fusionsfr{} vs ROCKET (blue), \fusionsr{} vs ROCKET (red). Annotated $n$ values indicate the number of datasets where the first model outperforms the second. Green annotations mark extreme F3 wins (e.g., \textit{RefrigerationDevices}, \textit{Rock})}
\label{fig:accuracy_gains}

\vspace{0.5cm}

\centering
\includegraphics[width=\textwidth]{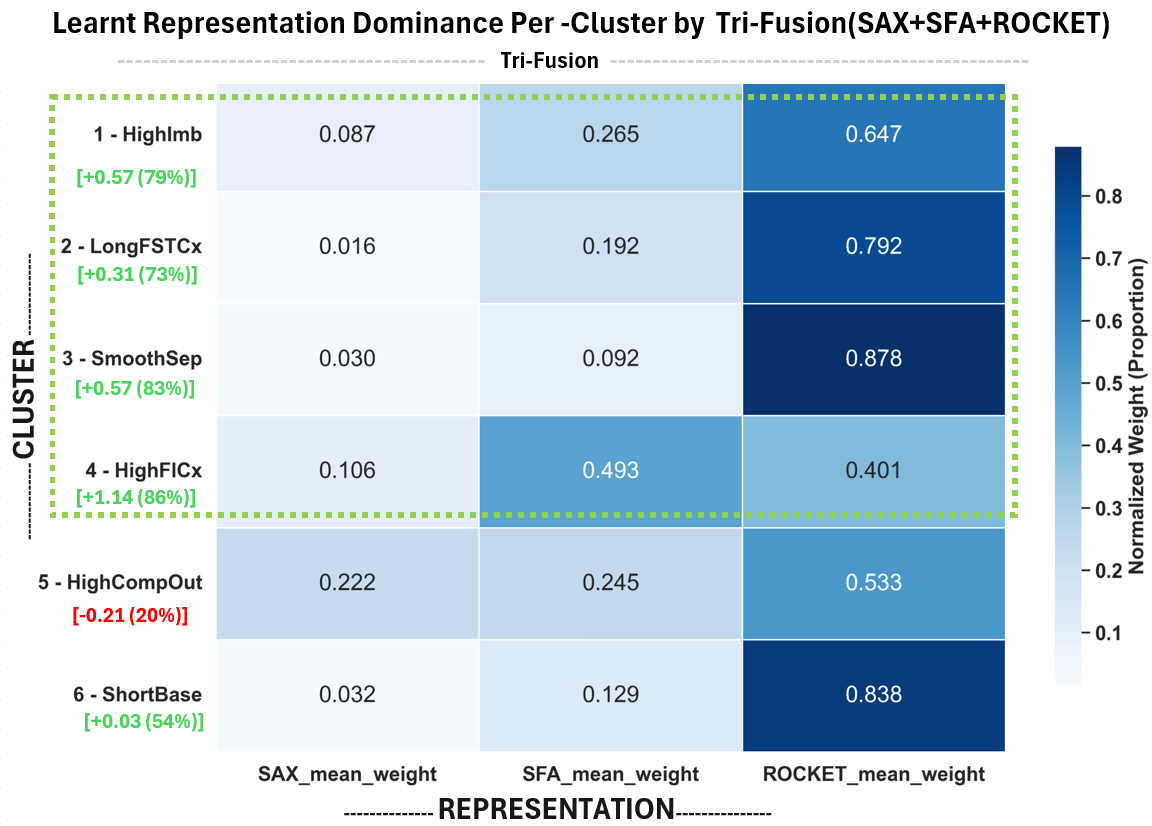}

\captionof{figure}{Normalized mean representation weights learned by the Tri-Fusion model. For each cluster, weights were first averaged across datasets and then normalized to sum to 1, ensuring interpretability as relative proportions of SAX, SFA, and ROCKET. \textbf{Annotations:} Values in brackets below cluster labels indicate the mean accuracy improvement (percentage points) and win rate (\%) of the fusion model relative to the ROCKET-only baseline. Green text denotes positive gains, while red text denotes negative performance. \textbf{Green dashed boxes} highlight clusters where the fusion model achieved both a positive accuracy gain and a win rate above 55\%.}
\label{fig:weight_dominance}
\end{samepage}

\FloatBarrier
\begin{samepage}
\subsubsection{D.4. SHAP Ablations}

\vspace{0.5cm}

\centering
\includegraphics[width=0.8\textwidth]{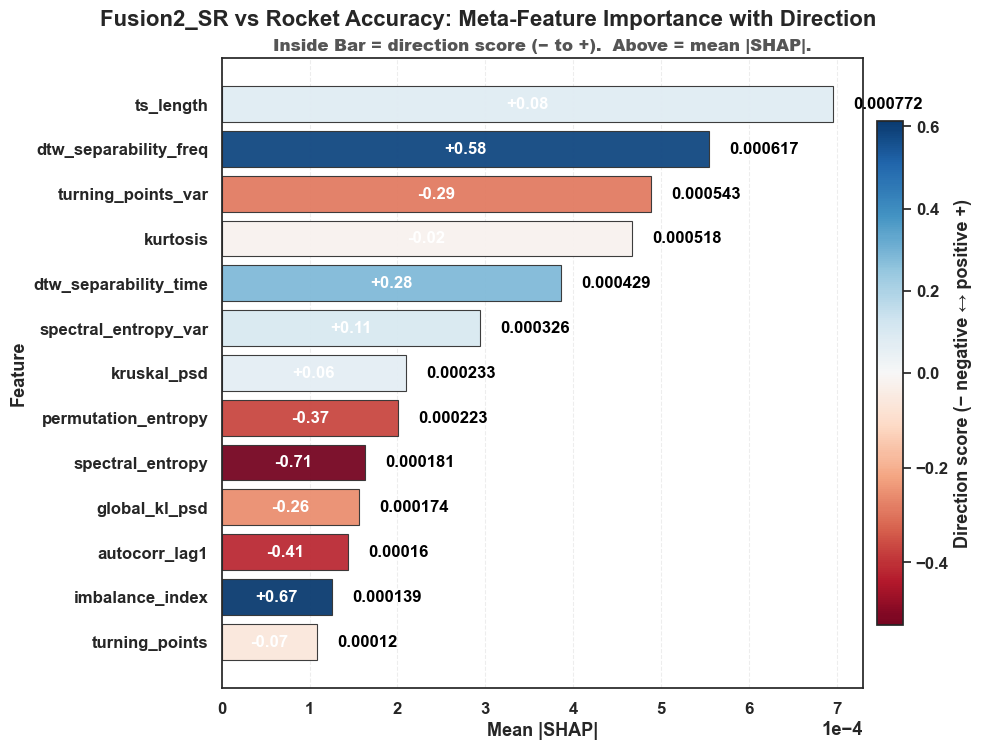}

\captionof{figure}{SHAP summary for predicting $\Delta\!\operatorname{acc}=\operatorname{acc}(\fusionsr{})-\operatorname{acc}(\rocket)$ from meta-features. Higher mean absolute SHAP indicates stronger global influence; signs are taken from the SHAP expectation.}
\label{fig:shap2_sr}

\vspace{0.2cm}

\centering
\includegraphics[width=0.8\textwidth]{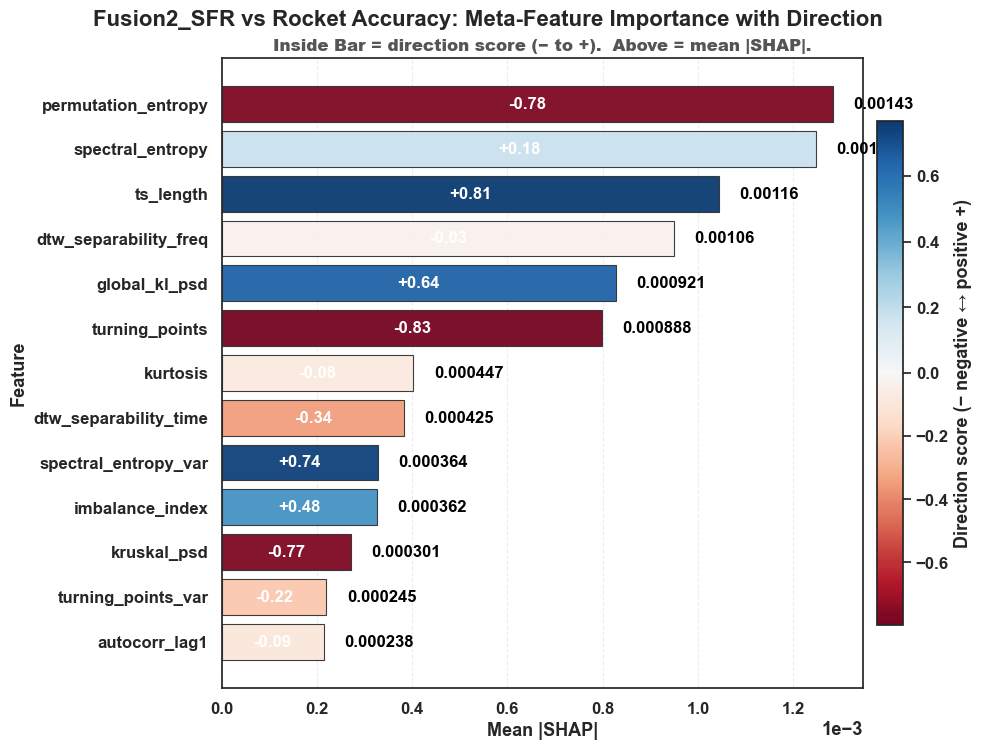}

\captionof{figure}{SHAP summary for predicting $\Delta\!\operatorname{acc}=\operatorname{acc}(\fusionsfr{})-\operatorname{acc}(\rocket)$ from meta-features. Higher mean absolute SHAP indicates stronger global influence; signs are taken from the SHAP expectation. Length remains amongst the top Fusion win predictors across both bi and tri fusions.}
\label{fig:shap2_sfr}

\vspace{0.5cm}

\centering
\includegraphics[width=\textwidth]{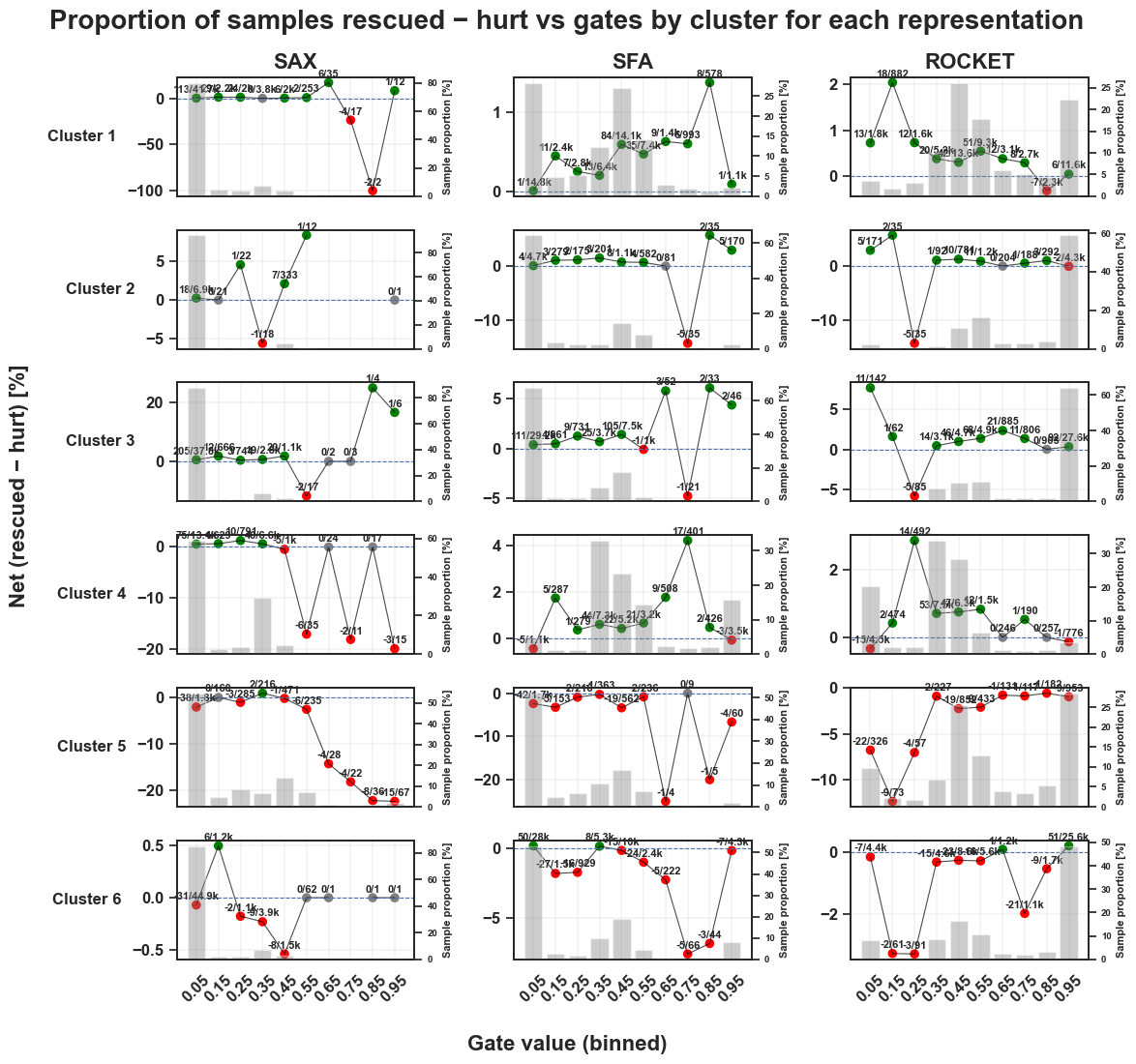}

\captionof{figure}{
    Cluster-wise relationship between gate values and fusion benefit,
    with sample proportions.
    Rows correspond to clusters (C1--C6) and columns to representations
    (SAX, SFA, ROCKET). In each panel, the black line shows net
    (rescued $-$ hurt) percentage as a function of the gate value; green
    (red) markers indicate positive (negative) net values. Light grey bars
    in the background show the proportion of samples falling into each gate
    bin. HighCompOut (C5) stands out: bins with substantial mass at high SFA
    and, to a lesser extent, high SAX exhibit strongly negative net
    rescued--hurt, whereas ROCKET-dominated bins remain close to neutral. Other clusters, e.g.\ C4, show that non--ROCKET-dominated
    regions can be neutral or beneficial, highlighting that the harmful
    regime is specific to C5 rather than a global property of SAX or SFA.
  }
\label{fig:gate-cluster-grid}
\end{samepage}

\FloatBarrier
\begin{samepage}
\subsubsection{D.5. Case Study: ROCK Dataset}
\label{app:rock-case}

\vspace{0.1cm}

\centering

\begin{minipage}{\linewidth}
  \centering
  \includegraphics[width=0.7\linewidth]{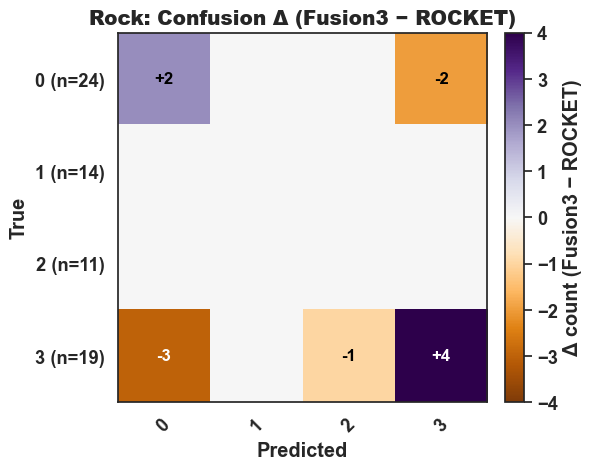}
  \vspace{0.1em}
  \par\noindent\textbf{(a)} Error structure and status breakdown.
  \label{fig:rock-matrix}
\end{minipage}

\vspace{0.1em}

\begin{minipage}{\linewidth}
  \centering
  \includegraphics[width=0.7\linewidth]{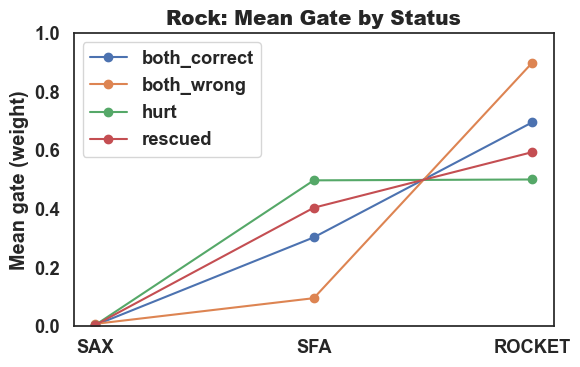}
  \vspace{0.1em}
  \par\noindent\textbf{(b)} Gating behaviour and error fixes.
  \label{fig:rock-gates}
\end{minipage}

\captionof{figure}{
    ROCK dataset case study: relation between gating behaviour and error fixes.
    \textbf{(a)} Matrix-style view of per-class performance and
    sample-level statuses (both-correct, rescued, hurt, both-wrong) under
    ROCKET and F3. This highlights which classes benefit most
    from fusion and where errors persist.
    \textbf{(b)} Corresponding gating behaviour: net
    (rescued $-$ hurt) percentage as a function of the SAX, SFA, and ROCKET
    gate values for ROCK, with light grey bars indicating the proportion of
    samples per bin. Error fixes (rescued samples) are concentrated in bins
    where the gate shifts modestly away from weak experts and towards ROCKET,
    while hurt cases occur when the gate under-weights ROCKET or over-emphasises
    less reliable representations. Together, these plots provide a concrete
    example of how dataset-level error patterns align with the learned gating
    policy on ROCK.
  }
\label{fig:rock-case}
\end{samepage}

\clearpage
\bibliographystyle{splncs04}
\bibliography{refs}

\end{document}